\documentclass[10pt,twocolumn,letterpaper]{article}

\usepackage[pagenumbers]{cvpr} 
\usepackage{multirow}
\usepackage{url}
\usepackage[T1]{fontenc}
\definecolor{cvprblue}{rgb}{0.21,0.49,0.74}
\usepackage[pagebackref,breaklinks,colorlinks,allcolors=cvprblue]{hyperref}

\definecolor{myPurple}{rgb}{0.4, .0, .8}
\definecolor{myGreen}{rgb}{0, .8, .3}
\definecolor{myRed}{rgb}{0.8, .2, .2}
\definecolor{myOrange}{rgb}{0.8, 0.45, 0.0}
\definecolor{myBlue}{rgb}{.0, .0, 1.0}

\newcommand\blfootnote[1]{%
  \begingroup
  \renewcommand\thefootnote{}\footnote{#1}%
  \addtocounter{footnote}{-1}%
  \endgroup
}

\def\paperID{ }
\def\confName{CVPR}
\def\confYear{2026}

\title{UIKA: Fast Universal Head Avatar from Pose-Free Images}

\author{
  Zijian Wu$^{1,2,*}$\quad
  Boyao Zhou$^{2,\#}$\quad
  Liangxiao Hu$^{2}$\quad
  Hongyu Liu$^{2,3}$\quad
  Yuan Sun$^{2,4}$\\
  Xuan Wang$^{2,4}$\quad
  Xun Cao$^{1}$\quad
  Yujun Shen$^{2}$\quad
  Hao Zhu$^{1,\dagger}$\\[6pt]
  \normalsize $^1$Nanjing University \quad
  \normalsize $^2$Ant Group \quad 
  \normalsize $^3$HKUST \quad
  \normalsize $^4$Xi'an Jiaotong University \\
}

\usepackage{booktabs}
\usepackage{bbding}
\usepackage{utfsym}
\usepackage{colortbl}
\usepackage{subcaption}
\usepackage{caption}

\usepackage{threeparttable}
\usepackage{pifont}
\usepackage{bm}
\usepackage{mathtools}

\begin{document}

\twocolumn[{%
\renewcommand\twocolumn[1][]{#1}%
\maketitle
\begin{center}
    \centering
    \captionsetup{type=figure}
    \vspace{-0.3in}
     \includegraphics[width=\textwidth]{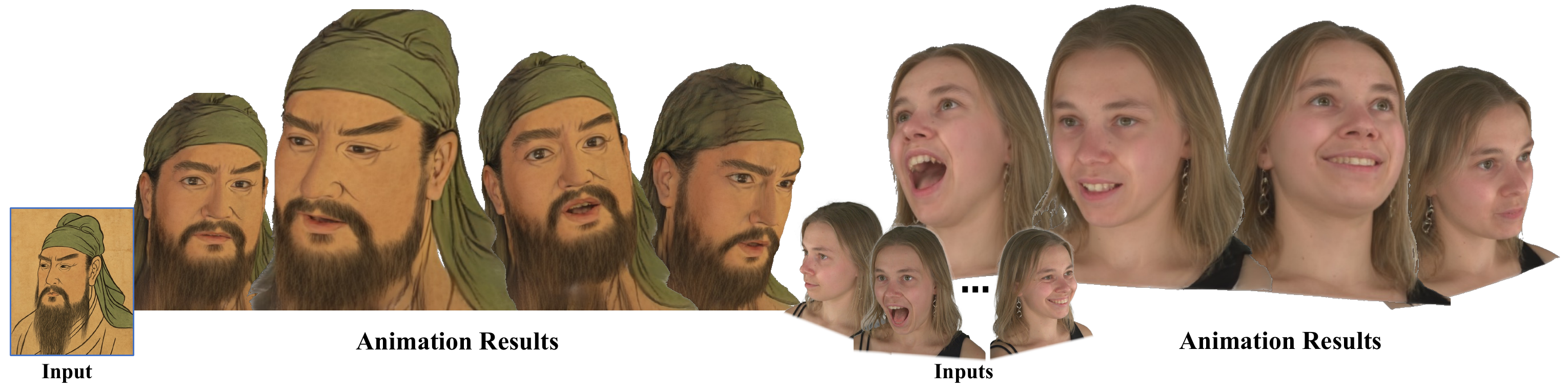}
     \vspace{-0.2in}
      \caption{We present \textsc{Uika}, a novel \textbf{feed-forward} approach for high-fidelity 3D Gaussian head avatar reconstruction from \textbf{an arbitrary number} of input images (e.g., \textit{a single portrait image} or \textit{multi-view captures}) \textbf{without requiring} extra camera or expression annotations.}
      \vspace{-0.0in}
      \label{fig:teaser}
\end{center}%
}]

\maketitle

\blfootnote{* Work done during an internship at Ant Group.}
\blfootnote{\# Project lead. $\dagger$ Corresponding author.}

\begin{abstract}

We present \textsc{UIKA}, a feed-forward animatable Gaussian head model from an arbitrary number of pose-free inputs, including a single image, multi-view captures, and smartphone-captured videos.
Unlike the traditional avatar method, which requires a studio-level multi-view capture system and reconstructs a human-specific model through a long-time optimization process, we rethink the task through the lenses of model representation, network design, and data preparation.
First, we introduce a UV-guided avatar modeling strategy, in which each input image is associated with a pixel-wise facial correspondence estimation.
Such correspondence estimation allows us to reproject each valid pixel color from screen space to UV space, which is independent of camera pose and character expression.
Furthermore, we design learnable UV tokens on which the attention mechanism can be applied at both the screen and UV levels.
The learned UV tokens can be decoded into canonical Gaussian attributes using aggregated UV information from all input views.
To train our large avatar model, we additionally prepare a large-scale, identity-rich synthetic training dataset.
Our method significantly outperforms existing approaches in both monocular and multi-view settings.
See more details in our project page: \small\url{https://zijian-wu.github.io/uika-page/}.

\end{abstract}
\section{Introduction}
\label{sec:intro}

Creating a 3D-aware human portrait avatar is a crucial research area for downstream applications such as tele-presence systems, the filmmaking industry, and virtual reality.
This area remains challenging in two ways: lifelike avatar quality and a flexible capture setup.
Our goal is to reconstruct a photo-realistic avatar model from an arbitrary number of pose-free images, eliminating the requirement for estimating camera and expression parameters.

Early 2D approaches~\cite{gong2023toontalker,wang2023progressive,xu2024vasa,zhang2023metaportrait,ma2024followyourpose,liu2023human} leverage the powerful generative capabilities of GANs~\cite{goodfellow2020generative, karras2019style, Karras2021} to drive source images by integrating facial landmarks~\cite{zakharov2019few, siarohin2019first} or latent codes~\cite{burkov2020neural} as control signals.
Recent methods~\cite{ma2024followyouremoji, ma2025followfaster, Gao2025Learn2Control, zhao2025xnemoexpressiveneuralmotion} leverage advances in diffusion models to improve animation performance further.
Although these 2D approaches achieve promising results, they often exhibit long inference times and are not robust to extreme camera poses due to the lack of explicit 3D representation.

In terms of 3D avatar modeling, classic methods~\cite{qian2024gaussianavatars, lee2025surfhead, xiang2024flashavatar, zhang2025fate} typically require long-term optimization for a specific identity using studio-level videos.
In particular, a sophisticated multi-view camera system is always necessary for comprehensive 3D reconstruction with the representation of either NeRF~\cite{mildenhall2021nerf} or Gaussian-Splatting~\cite{Xu_2024_CVPR,qian2024gaussianavatars,cai2025hera,wang20253d}.
Such methods demand accurate camera calibration, while some approaches~\cite{Xu_2024_CVPR,liao2025hhavatar} rely on computationally intensive post-processing networks, thereby hindering their practical deployment in downstream applications.
Even if some approaches~\cite{chen2024monogaussianavatar} take monocular videos as input, they typically rely on high-precision expression capture data to ensure accurate avatar modeling.
Regardless of whether monocular or multi-view data are used as input, such optimization-based methods remain fundamentally constrained in generalizing to novel portrait expressions and camera poses.

Recently, emerging methods have shifted towards feed-forward avatar modeling by leveraging a large reconstruction model~\cite{hong2023lrm,xu2024grm,wang2024pf, Sketch2PoseNet2025wang} from a single image~\cite{chu2024gagavatar,he2025lam} or limited images~\cite{zhao2024invertavatar,kirschstein2025avat3r}.
LAM~\cite{he2025lam} and GAGAvatar~\cite{chu2024gagavatar} reconstruct head avatars from a single input image and are typically trained on monocular portrait videos, which often limits their ability to generalize to novel view synthesis under large camera poses.
Avat3r~\cite{kirschstein2025avat3r}, in contrast, requires a fixed set of four calibrated input images, a restrictive setting that reduces practical applicability and also confines training to existing identity-scarce multi-view datasets, thereby limiting generalization.
More recent methods, such as GPAvatar~\cite{chu2024gpavatar} and PF-LHM~\cite{qiu2025pf}, extend the input setting to an arbitrary number of images, but they lack explicit correspondence across input frames, making multi-view information aggregation less reliable. Tab.~\ref{tab:comp_settings} summarizes the flexibility and efficiency of our method relative to the baselines.

\begin{table}[]
\centering
\scalebox{0.9}{\begin{tabular}{cccccc}
\toprule
\textbf{Method} & \textbf{Inputs}  & \textbf{FF} & \textbf{PF} & \textbf{RTA} \\ \hline
GAGAvatar~\cite{chu2024gagavatar}          & 1          & \ding{51}   & \ding{51} & \ding{51} \\
Portrait4D-v2~\cite{deng2024portrait4dv2}          & 1          & \ding{51}  & \ding{55} & \ding{55} \\
AvatarArtist~\cite{liu2025avatarartist}          & 1          & \ding{51}   & \ding{51} & \ding{55} \\
LAM~\cite{he2025lam}      & 1          & \ding{51}   & \ding{51} & \ding{51} \\
FastAvatar~\cite{liang2025fastavatar}      & 1          & \ding{51}   & \ding{51} & \ding{51} \\
Avat3r~\cite{kirschstein2025avat3r}        & 4          & \ding{51}   & \ding{55}  & \ding{55} \\
CAP4D~\cite{taubner2025cap4d}        & $\geq$ 1          & \ding{55}   & \ding{55}  & \ding{51} \\
GPAvatar~\cite{chu2024gpavatar}        & $\geq$ 1          & \ding{51}   & \ding{51}  & \ding{55} \\
InvertAvatar~\cite{zhao2024invertavatar}        & $\geq$ 1          & \ding{55}   & \ding{55}  & \ding{55} \\
DiffusionRig~\cite{ding2023diffusionrig}        & $\geq$ 1          & \ding{55}   & \ding{55}  & \ding{55} \\
FastAvatar~\cite{wu2026fastavatar}        & $\geq$ 1          & \ding{51}   & \ding{55}  & \ding{51} \\
Ours         & $\geq$ 1 &  \ding{51} & \ding{51} & \ding{51} \\ \bottomrule
\end{tabular}}
\vspace{-0.1in}
\caption{\textbf{Comparison with state-of-the-art 3D head reconstruction methods.} \textbf{FF} denotes a feed-forward pipeline that requires no test-time optimization or fine-tuning. \textbf{PF} indicates pose-free input, i.e., camera and expression labels are not required. \textbf{RTA} denotes real-time animatability ($\geq$ 30 FPS).}
\label{tab:comp_settings}
\vspace{-15px}
\end{table}

In this work, we present \textsc{UIKA}, a novel feed-forward framework for animatable Gaussian head modeling from an arbitrary number of pose-free input images, as shown in Fig.~\ref{fig:teaser}.
To establish explicit correspondences across pose-free input images, we design a facial correspondence estimator that supports an arbitrary number of inputs, inspired by Pixel3DMM~\cite{giebenhain2026pixeldmm}.
Given a set of pose-free input images, our facial correspondence estimator first estimates UV coordinates in the pixel level, and the corresponding colors are reprojected onto the shared UV space.
The reprojected images and original images are embedded with a frozen DINOv3~\cite{simeoni2025dinov3} encoder followed by a trainable lightweight fusion module, producing multi-scale features from both screen and UV spaces.
Typically, prior works~\cite{zhuang2024idolinstant, he2025lam, Qiu_2025_LHM, qiu2025pf} build a connection between learnable tokens and screen space features by using Transformer blocks, which lack a structural correspondence. 
Other than conventional screen attention, we introduce a UV attention branch that enables our learnable UV tokens to interact with UV-space features.
This design allows our model to simultaneously leverage local details from the screen space and structured global context from the reprojected UV space in a complementary manner.
Furthermore, the processed learnable tokens can be decoded into canonical Gaussian primitives, including appearance and geometry attributes.
Although the predicted appearance is globally coherent to input images, it typically lacks realistic details. 
Thus, we propose a self-adaptive fusion strategy per Gaussian primitive that blends these two color sources via learned weights.
This design dynamically balances accurate but potentially incomplete local cues against globally coherent yet sometimes imprecise predictions, leading to high reconstruction quality.
In addition, the resulting canonical Gaussian head avatar is immediately animatable using standard linear blend skinning and supports real-time rendering at 220 FPS, in contrast to approaches~\cite{chu2024gagavatar, chu2024gpavatar, oroz2025percheadperceptualheadmodel} that rely on an additional neural renderer at inference time to produce the final outputs.
To strengthen multi-view learning, we construct a synthetic dataset with diverse identities and rich expression variations.
Training on our collected and synthetic datasets, our method outperforms the prior state of the art in both monocular and multi-view settings.
The contributions of our work can be summarized as:
\begin{itemize}

\item We present \textsc{UIKA}, a feed-forward framework that reconstructs animatable 3D Gaussian head avatars from an arbitrary number of pose-free input images.

\item We design a novel UV attention branch that leverages predicted facial correspondence to efficiently align multi-observation within a unified canonical space, facilitating robust cross-image information matching.

\item We propose a self-adaptive fusion strategy to dynamically balance predicted global appearance and reprojected local details from input images to improve overall quality.

\item To mitigate the limited identity diversity, view coverage, and motion in existing datasets, we curate a large-scale, multi-view synthetic head dataset for head avatar reconstruction and generation.

\end{itemize}

\section{Related Work}
\label{sec:related}

\subsection{Generative 2D Head Avatar}
\label{subsec:2dhead}

Early one-shot 2D methods are typically built upon conditional GANs~\cite{gal2022stylegan,gfpgan,goodfellow2020generative, karras2019style, Karras2021,zhao2024invertavatar,sun2023next3d} and explicit warping~\cite{zakharov2019few,siarohin2019first,hong2022depth,burkov2020neural,siarohin2019animating,siarohin2021motion,wang2023progressive,deng2024portrait4d, deng2024portrait4dv2}, where a motion representation is estimated to deform a static reference image. More recently, diffusion models~\cite{xie2024x, ma2024followyouremoji, wei2024aniportrait, ding2023diffusionrig, shi2025dex, wang2025tera} have substantially improved fidelity and robustness in one-shot portrait animation, benefiting from scalable architecture (e.g., DiT~\cite{Peebles2022DiT, sd3mmdit, mir2023dithead}) and strong priors learned from large-scale data. Diffusion-based talking-head synthesis typically falls into two lines: audio-driven models~\cite{stypulkowski2024diffused,xu2024vasa,yu2023thpad} that generate speech-conditioned facial motion with realistic articulation and landmark-controlled models~\cite{wei2024aniportrait,ma2024followyouremoji,ma2025followfaster} that use sparse keypoint constraints for controllable reenactment and cross-domain transfer. Despite rapid progress, 2D diffusion-based methods such as DiffusionRig~\cite{ding2023diffusionrig} remain image-space with weakly 3D-aware performance, which often leads to limited extrapolation to extreme viewpoints, constrained long-range motion consistency, and high inference cost.

\subsection{Optimization-based 3D Head Avatar}
\label{subsec:3d-optim}

Optimization-based methods typically require long-time optimization for a specific person,
spanning mesh-centric pipelines~\cite{grassal2022nha}, volumetric neural fields~\cite{gafni2021nerface, hong2022headnerf, zielonka2023insta, mildenhall2021nerf, xu2023avatarmav, yu2024one2avatar}, and more recently animatable 3D Gaussians~\cite{xu2023gaussianheadavatar, xiang2024flashavatar, chen2024monogaussianavatar, xu2025gphm, zheng2024headgap, zhang2025fate}.
Recently, generative models have been introduced as an auxiliary source of supervision. CAP4D~\cite{taubner2025cap4d} and MVP4D~\cite{taubner2025mvp4d} leverage morphable multi-view diffusion to synthesize pseudo-observation supervision from reference images to regularize identity preservation and view completeness during reconstruction. Such hybrid pipelines narrow the gap between sparse-view inputs and multi-view reconstruction quality, but they rely on iterative fitting process and thus are still ill-suited for real-time or truly one-shot applications, motivating feed-forward 3D head avatar reconstruction.

\subsection{Feed-Forward 3D Head Avatar}
\label{subsec:3d-ff}

Feed-forward 3D head avatar methods orient to a single network forward pass reconstruction by learning strong identity priors from large-scale datasets, including in-the-wild monocular videos~\cite{xie2022vfhq, zhang2021flow}, studio-quality multi-view captures~\cite{kirschstein2023nersemble, ava256, pan2024renderme360, yang2020facescape, buehler2024cafca} and synthetic data~\cite{deng2019accurate,feng2021learning,zhang2025bringingportrait3dpresence,deng2024portrait4d,deng2024portrait4dv2,liu2025avatarartist}. Early approaches~\cite{chan2022efficient, yu2023nofa, li2023one, li2024generalizable, ma2023otavatar, zhuang2022mofanerf, trevithick2023real, ye2024real3d, yu2025realityavatar} can infer a personalized, animatable avatar from one or a handful of images. However, the NeRF-based~\cite{mildenhall2020nerf} method, such as GPAvatar~\cite{chu2024gpavatar}, struggles to support real-time avatar animation due to implicit modeling. Recent works~\cite{liang2025fastavatar, zhao2026generalizable, oroz2025percheadperceptualheadmodel, peng2025flexavatar, kirschstein2025flexavatar, ji2026fastgha} adopt 3D Gaussian Splatting~\cite{kerbl3Dgaussians} as a representation to enable real-time and high-fidelity rendering. For example, LAM~\cite{he2025lam} and GAGAvatar~\cite{chu2024gagavatar} can reconstruct a 3D head avatar through a single portrait input, but often degrade under large viewpoint extrapolation for unobserved regions. More recently, Avat3r~\cite{kirschstein2025avat3r} and HeadGAP~\cite{zheng2024headgap} extend to multiple inputs to improve 3D consistency. However, they are trained totally on multi-view datasets with limited identity diversity, which can hinder generalization to in-the-wild inputs. FastAvatar~\cite{wu2026fastavatar} accepts an arbitrary number of inputs but suffers from redundant Gaussian points when using more and more views. In contrast, our method reconstructs high-fidelity and animatable head avatars in a single feed-forward pass from any number of images without requiring additional camera or expression annotations and supports joint training on identity-rich monocular videos and 3D-consistent multi-view data for efficient and scalable deployment.

\section{Method}
\label{sec:method}

\begin{figure*}[htbp!]
    \centering
    \includegraphics[width=\textwidth]{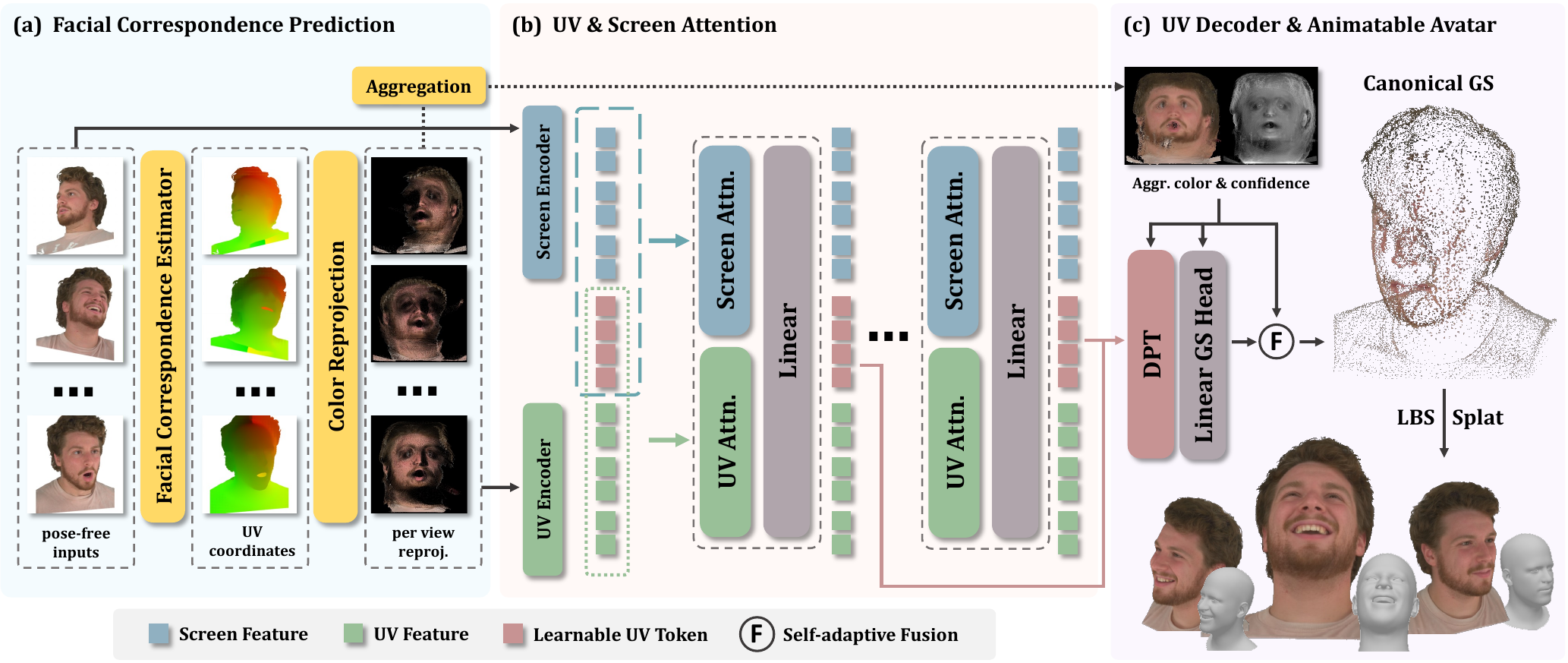}
    \caption{\textbf{Pipeline Overview.} Given a set of pose-free input images, our pipeline begins with a facial correspondence estimator that predicts UV coordinates for valid facial pixels, and the corresponding colors are reprojected onto the shared UV space. The source images (screen space) and reprojected images (UV space) are encoded through two dedicated encoders, producing multi-scale features from both screen space and UV space. We then apply screen attention and UV attention to inject these into learnable UV tokens, which are then decoded into UV Gaussian attribute maps while incorporating the aggregated color and confidence map. The resulting canonical Gaussian head avatar supports animation via standard linear blend skinning and achieves real-time rendering at 220 FPS.}
    \label{fig: pipeline}
    \vspace{-15px}
\end{figure*}

Given an arbitrary number of pose-free images $\{\mathrm{I}^i_{\text{s}}\}^N_{i=1}$, without additional camera or expression parameters, our goal is to reconstruct a high-fidelity and animatable Gaussian head avatar represented by a set of Gaussians~\cite{kerbl3Dgaussians} $\mathcal{G} = \{ \textbf{c}_k, o_k, \bm{\mu}_k, \bm{s}_k, \bm{r}_k \}^M_{k=1}$. The overall pipeline is illustrated in Fig.~\ref{fig: pipeline}.
Firstly, we present a facial correspondence estimator and introduce the color reprojection and aggregation in Sec.~\ref{subsec:uv_pred}. 
Then, Sec.~\ref{subsec:dual_attn} introduces a novel UV attention branch into Transformer architecture. In the Sec.~\ref{subsec:uv_dec}, we present the proposed self-adaptive fusion strategy in the UV decoder. 
We furthermore introduce our synthetic multi-view head dataset in Sec.~\ref{subsec:synth}. 
Finally, Sec.~\ref{subsec:loss} outlines the training objectives.

\subsection{Facial Correspondence Prediction}
\label{subsec:uv_pred}

\textbf{Facial correspondence estimator.} Inspired by prior work~\cite{giebenhain2026pixeldmm, wang2025vggt}, we develop a facial correspondence estimator that accepts an arbitrary number of pose-free images $\{\mathrm{I}^i_{\text{s}}\}^N_{i=1}$ as input and predicts facial correspondence, in the format of pixel-aligned UV coordinates $\{\mathrm{U}^i\}^N_{i=1}, \quad \mathrm{U} = (u, v) \in [0, 1]^2$, for corresponding input images as follow:
\begin{align}
\mathrm{U}^i = \mathcal{U}\left(\mathrm{I}^i_{\text{s}}\right),  \quad i \in [1, N];
\end{align}
where $\mathcal{U} \ (\cdot)$ denotes our facial correspondence estimator network. Specifically, the input images are processed with a frozen pre-trained encoder~\cite{simeoni2025dinov3}, which extracts robust feature representations. These features are subsequently decoded into dense UV coordinate maps through a trainable DPT head~\cite{Ranftl2020dpt2, Ranftl2021dpt1}. Further architectural details are provided in the supplementary material.

\noindent\textbf{Color reprojection \& Aggregation.} As shown in Fig.~\ref{fig: pipeline} (a), we reproject input images $\{\mathrm{I}^i_{\text{s}}\}^N_{i=1}$ from screen-space into a shared UV space by leveraging the predicted facial correspondence $\{\mathrm{U}^i\}^N_{i=1}$, alleviating the ambiguity of camera pose and facial expression from different frames. Thus, we can obtain reprojected images $\{\mathrm{I}^i_{\text{uv}}\}^N_{i=1}$ by pixel-to-pixel matching. We then aggregate all reprojected images into an averaged UV observation $\mathrm{I}_{\text{aggr}}$ and a confidence map $\gamma_{\text{aggr}}$:
\begin{align}
\mathrm{I}^i_{\text{uv}} = \ \mathrm{Reproj} \left( \mathrm{I}^i_{\text{s}}, \mathrm{U}^i \right),  \quad i \in [1, N]; \\ 
\mathrm{I}_{\text{aggr}}, \ \gamma_{\text{aggr}} \leftarrow \mathrm{Aggr} \left( \mathrm{I}^1_{\text{uv}},\mathrm{I}^2_{\text{uv}},\dots,\mathrm{I}^N_{\text{uv}} \right);
\end{align}
In practice, $\mathrm{I}_{\text{aggr}}$ is computed by pixel-wise averaging over the reprojected images. For each UV pixel, we count the number of valid projections $n_{\text{hit}}$ and define the aggregated confidence as $\gamma_{\text{aggr}} \coloneqq \operatorname{Norm}\left(\log(1+n_{\text{hit}})\right)$, here $\operatorname{Norm} \ (\cdot)$ denotes \emph{min–max normalization}.

\subsection{UV \& Screen Attention}
\label{subsec:dual_attn}

As shown in Fig.~\ref{fig: pipeline} (b), given source input images $\{\mathrm{I}^i_{\text{s}}\}^N_{i=1}$ and their corresponding reprojected images $\{\mathrm{I}^i_{\text{uv}}\}^N_{i=1}$, we extract screen features $ \mathcal{F}_{\text{s}} $ and UV features $ \mathcal{F}_{\text{uv}} $ via:
\begin{align}
\mathcal{F}_{\text{j}} &= \mathcal{E}_{\text{j}} \left( \mathrm{I}^1_{\text{j}} \right) \oplus \mathcal{E}_{\text{j}} \left( \mathrm{I}^2_{\text{j}} \right) \oplus \dots \oplus \mathcal{E}_{\text{j}} \left( \mathrm{I}^N_{\text{j}} \right);
\end{align}
where $\text{j} \in [\text{s}, \text{uv}]$ for either screen or UV space and $\oplus$ denotes the concatenation in the length dimension. Encoder $\mathcal{E}_{\text{j}}$ is composed of a frozen pretrained DINOv3~\cite{simeoni2025dinov3} backbone and a trainable lightweight CNN fusing features derived from both shallow and deep layers of the backbones.

To exploit semantic features from both screen space and UV space into our learnable UV tokens $\mathcal{Z} \in \mathbb{R}^{\mathrm{L}_{\text{z}} \times \mathrm{D}}$, we perform attention mechanism~\cite{sd3mmdit} $\mathrm{Attn}$ in both spaces:
\begin{align}
\Delta \mathcal{Z}_{\text{j}},  \ \Delta \mathcal{F}_{\text{j}} &= \mathrm{Attn}_{\text{j}} \left( \mathcal{Z}, \mathcal{F}_{\text{j}} \right); \\
\mathcal{Z}^{\prime} &= \mathcal{Z} + \mathrm{ML} \mathrm{P} \left( \mathcal{Z} + \Delta \mathcal{Z}_{\text{s}} + \Delta \mathcal{Z}_{\text{uv}} \right); \\
\mathcal{F}^{\prime}_{\text{j}} &= \mathcal{F}_{\text{j}} + \mathrm{MLP} \left( \mathcal{F}_{\text{j}} + \Delta \mathcal{F}_{\text{j}} \right);
\end{align}
where $\text{j} \in [\text{s}, \text{uv}]$ for either screen or UV space and $\mathcal{Z}^{\prime}, \mathcal{F}^{\prime}_{\text{j}}$ denotes the updated $\mathcal{Z}, \mathcal{F}_{\text{j}}$ in a Transformer block.

\subsection{UV Decoder}
\label{subsec:uv_dec}

\textbf{UV Gaussian prediction.} As shown in Fig.~\ref{fig: pipeline} (c), starting from the UV aggregation map $\{ \mathrm{I}_{\text{aggr}}, \gamma_{\text{aggr}} \}$ produced in Sec.~\ref{subsec:uv_pred} and the multi-depth learned UV tokens $\mathcal{Z}^l$ obtained from our Transformer in Sec.~\ref{subsec:dual_attn}, we feed them into our UV decoder $\mathcal{D} \ (\cdot)$ to obtain the canonical Gaussian attributes:
\begin{align}
\{ \hat{\bm{c}}_k, w_k, o_k, \bm{\Delta\mu}_k, \bm{s}_k, \bm{r}_k \}^M_{k=1} = \mathcal{D} \left( \mathcal{Z}^l; \mathrm{I}_{\text{aggr}}, \gamma_{\text{aggr}} \right);
\end{align}
where $l = 3,6,9,12$ denotes different depth of our Transformer blocks and $\hat{\bm{c}}_k, w_k, o_k, \bm{\Delta\mu}_k, \bm{s}_k, \bm{r}_k$ represent the predicted color, color fuse weight, opacity, position offset, scaling, and rotation of the canonical Gaussian attributes, respectively. 
And we define a \textbf{self-adaptive fusion strategy} to balance the impact between the predicted $\bm{c}_k$ and real-captured aggregation $\bm{c}_k^{aggr}$ via:
\begin{align}
\bm{c}_k = w_k * \ &\hat{\bm{c}}_k + (1 - w_k) * \bm{c}_k^{aggr}, \quad \bm{c}_k^{aggr} \subset \mathrm{I}_{aggr};
\end{align}
The final canonical Gaussians $\mathcal{G}$ is updated with color $\bm{c}_k$ and position $\bm{\mu}^{m}_k + \bm{\Delta\mu}_k$, where $\bm{\mu}^{m}_k$ represents the initial position on the template FLAME~\cite{sigasia17FLAME} mesh surface.

\noindent\textbf{Novel expression animation.} Given the reconstructed canonical Gaussian head avatar, we reenact it under novel FLAME poses and expressions. Each Gaussian originates from a valid UV pixel; FLAME UV rasterization~\cite{xiang2024flashavatar, zhang2025fate} provides its associated triangle assignments and corresponding barycentric coordinates. Using barycentric interpolation over the corresponding mesh triangle, we obtain per-Gaussian quantities, e.g., LBS weights, posedirs, and shapedirs. Conditioned on target FLAME pose and expression, we then apply standard vertex-based linear blend skinning (LBS) to deform the Gaussians from the canonical space to the posed space, yielding the animated head avatar. Finally, we obtain the rendered images $\mathrm{I}_{\text{pred}}$ through differentiable Gaussian splatting $\mathcal{R} \ (\cdot)$ as follows:
\begin{align}
\mathrm{I}_{\text{pred}} = \mathcal{R} \left( \mathrm{LBS} \left(\mathcal{G}, \ \Theta  \right), \ \Pi \right);
\end{align}
where $\Theta$ denotes the target FLAME pose and expression parameters and $\Pi$ denotes the target camera parameters.

\subsection{Synthetic Multi-view Head Dataset Curation}
\label{subsec:synth}

Prior work predominantly trains on head datasets that are monocular, leading to restricted camera viewpoints and limited expression variability dominated by speech-related motions. Although recent multi-view datasets such as NeRSemble~\cite{kirschstein2023nersemble}, Ava-256~\cite{ava256}, and RenderMe-360~\cite{pan2024renderme360} alleviate the view limitation, they suffer from small identity counts due to costly capture setups and are typically recorded under studio lighting, hindering generalization to in-the-wild conditions. To address these limitations, we introduce a scalable data curation pipeline that combines a 3D head generation model with an efficient 2D portrait animation model to produce identity-diverse, multi-view sequences with extreme expressions. Concretely, we leverage SphereHead~\cite{li2024spherehead}, a 3D head generator trained on in-the-wild images spanning wide camera poses to synthesize multi-view and 3D consistent head renderings. For each identity, we sample 9 fixed viewpoints and render the corresponding views. We then employ LivePortrait~\cite{guo2024liveportrait}, an efficient 2D portrait animation model that drives a source head image using a driver video. For each view, we select the same driver sequence from a curated motion library to animate the rendered view, producing temporally synchronized multi-view head sequences. In total, we curate over 7,500 identities, each with 9 views and more than 13,000 frames per identity, covering complex and exaggerated facial expressions while avoiding expensive studio capture and improving robustness to in-the-wild scenarios. Further dataset details are provided in the supplementary material.

\subsection{Training Objectives}
\label{subsec:loss}

During training, we randomly sample 1 to $\mathrm{N}_{\text{ref}}$ frames from the same video as source inputs to reconstruct the canonical Gaussian representation, and additionally sample $\mathrm{N}_{\text{d}}$ frames as driving and target views for reenactment supervision. We supervise the rendered images against the corresponding ground truth frames using a photometric objective that combines L1, SSIM, and VGG-based perceptual losses: 
\begin{align}
\mathcal{L}_{\text{l1}} &= || \mathrm{I}_{\text{pred}} - \mathrm{I}_{\text{gt}} ||_1; \\
\mathcal{L}_{\text{lpips}} &= \mathrm{LPIPS} \left( \mathrm{I}_{\text{pred}}, \ \mathrm{I}_{\text{gt}} \right); \\
\mathcal{L}_{\text{ssim}} &= \mathrm{SSIM} \left( \mathrm{I}_{\text{pred}}, \ \mathrm{I}_{\text{gt}} \right);
\end{align}
We also add geometry regularization to prevent Gaussians from drifting too far from its initialized position via:
\begin{align}
\mathcal{L}_{\text{reg}} = || \max \left( \bm{\Delta\mu}, \ \epsilon \right) ||_2;
\end{align}
where $\bm{\Delta\mu}$ is the predicted Gaussian offsets, $\epsilon$ is an hyper parameters set close to 0. The overall training objectives are the weighted sum of the image supervision over all supervised frames and the offset regularization:
\begin{align}
\mathcal{L} = \lambda_{\text{l1}}\mathcal{L}_{\text{l1}} + \lambda_{\text{lpips}}\mathcal{L}_{\text{lpips}} + \lambda_{\text{ssim}}\mathcal{L}_{\text{ssim}} + \lambda_{\text{reg}} \mathcal{L}_{\text{reg}};
\end{align}
where $\lambda_{\text{l1}}$ and $\lambda_{\text{lpips}}$ are 1.0, $\lambda_{\text{ssim}}$ and $\lambda_{\text{reg}}$ are 0.1.

\section{Experiments}
\label{sec:exp}

\begin{figure*}[t]
    \centering
    \includegraphics[width=1.0\textwidth]{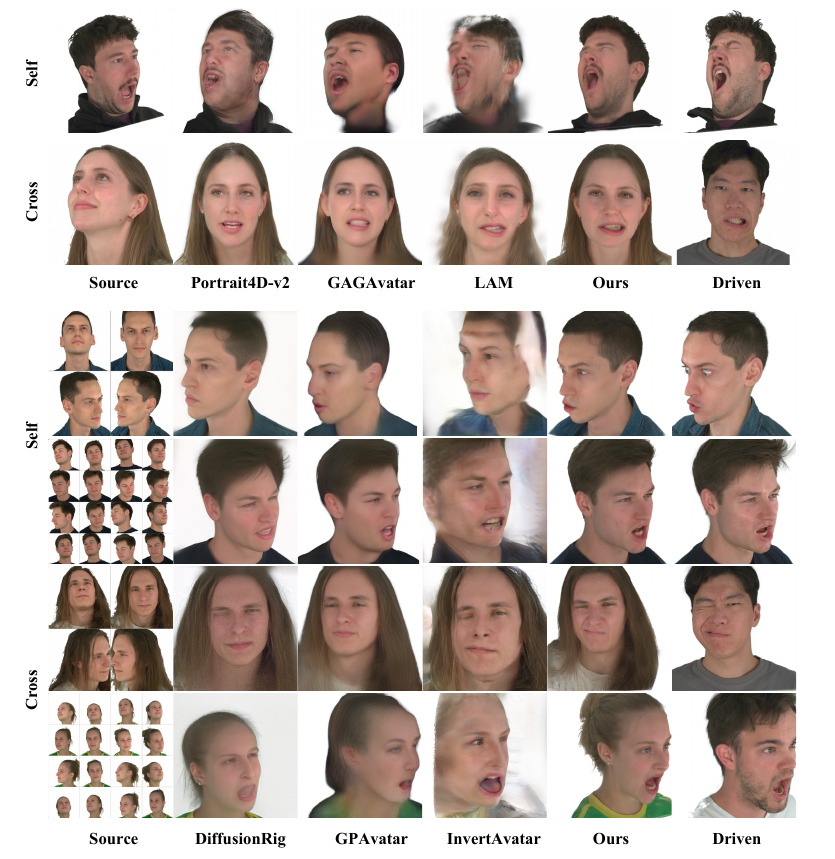}
    \caption{\textbf{Qualitative results for comparison to baselines in both monocular and multi-view settings in NeRSemble-v2 datasets.} 
    }
    \label{fig:comparison}
\end{figure*}

\begin{table*}[t]
\centering
\scalebox{1.0}{
\begin{tabular}{c|ccccccc|ccc}
\toprule
\multirow{2}{*}{\textbf{Method}} & \multicolumn{7}{c|}{\textbf{Self Reenactment}}  & \multicolumn{3}{c}{\textbf{Cross Reenactment}} \\
& \textbf{PSNR$\uparrow$}  & \textbf{SSIM$\uparrow$} & \textbf{LPIPS$\downarrow$} & \textbf{CSIM$\uparrow$}  & \textbf{AED$\downarrow$} & \textbf{APD$\downarrow$} & \textbf{AKD$\downarrow$} & \textbf{CSIM$\uparrow$} & \textbf{AED$\downarrow$} & \textbf{APD$\downarrow$}\\ \hline
Portrait4D-v2~\cite{deng2024portrait4dv2}     & 21.03   & 0.859   & 0.134  & 0.688 & 0.094   & 0.113   & 3.718  & 0.654 & 0.132   & 0.149 \\
GAGAvatar~\cite{chu2024gagavatar}         & 20.34   & 0.850   & 0.160  & 0.693 & 0.071   & 0.075   & 4.372  & \textbf{0.678} & 0.151   & 0.142 \\
LAM~\cite{he2025lam}               & 18.29   & 0.810   & 0.206  & 0.602 & 0.104   & 0.112   & 4.631  & 0.612 & 0.126   & 0.130 \\ \hline
Ours              & \textbf{21.69}   & \textbf{0.867}   & \textbf{0.105}  & \textbf{0.738} & \textbf{0.055}   & \textbf{0.056}   & \textbf{3.066}  & 0.649 & \textbf{0.114}   & \textbf{0.123} \\
\bottomrule
\end{tabular}
}
\vspace{-0.1in}
\caption{\textbf{Quantitative results on the monocular setting in VFHQ and NeRSemble-v2 datasets.}}
\label{tab:comp_metrics1}
\vspace{-0.15in}
\end{table*}

\begin{table*}[t]
\centering
\scalebox{1.0}{
\begin{tabular}{c|ccccccc|ccc}
\toprule
\multirow{2}{*}{\textbf{Method}} & \multicolumn{7}{c|}{\textbf{Self Reenactment}} & \multicolumn{3}{c}{\textbf{Cross Reenactment}} \\
& \textbf{PSNR$\uparrow$}  & \textbf{SSIM$\uparrow$} & \textbf{LPIPS$\downarrow$} & \textbf{CSIM$\uparrow$}  & \textbf{AED$\downarrow$} & \textbf{APD$\downarrow$} & \textbf{AKD$\downarrow$} & \textbf{CSIM$\uparrow$} & \textbf{AED$\downarrow$} & \textbf{APD$\downarrow$}\\ \hline
DiffusionRig~\cite{ding2023diffusionrig}      & 16.97   & 0.768   & 0.395  & 0.598 & 0.209   & 0.138   & 9.585  & 0.616 & 0.263   & 0.218 \\
GPAvatar~\cite{chu2024gpavatar}          & 17.11   & 0.783   & 0.313  & 0.553 & 0.129   & 0.108   & 6.423  & 0.492 & 0.210   & 0.168 \\
InvertAvatar~\cite{zhao2024invertavatar}      & 16.35   & 0.776   & 0.394  & 0.449 & 0.084   & 0.069   & 7.402  & 0.491 & 0.198   & 0.177 \\ \hline
Ours              & \textbf{22.50}   & \textbf{0.855}   & \textbf{0.120}  & \textbf{0.740} & \textbf{0.064}   & \textbf{0.063}   & \textbf{3.437}  & \textbf{0.666} & \textbf{0.145}   & \textbf{0.153} \\
\bottomrule
\end{tabular}
}
\vspace{-0.1in}
\caption{\textbf{Quantitative results on the multi-view setting in NeRSemble-v2 datasets.}}
\label{tab:comp_metrics2}
\vspace{-0.15in}
\end{table*}

\subsection{Experiments settings}
\label{subsec:settings}

\textbf{Implementation Details.} We implement our framework using PyTorch~\cite{pytorch}. Our Transformer architecture consists of $\mathrm{L} = 12$ MM-Transformer~\cite{sd3mmdit} blocks, each equipped with $\mathrm{h} = 16$ attention heads and a hidden feature dimension $\mathrm{D} = 1024$. For per-view reprojected images, the resolution is kept identical to the source inputs in $512 \times 512$. We introduce learnable UV tokens of length $ \mathrm{L}_{\text{z}} = 9216 $, which are then reshaped into a $ 96 \times 96 $ grid before being fed into the UV decoder. These UV tokens are jointly processed with the UV aggregate map via a DPT-based~\cite{Ranftl2021dpt1, Ranftl2020dpt2} decoder, producing a UV space feature map of size $ 384 \times 384 \times 256 $. Subsequently, we rasterize the UV representation of the FLAME~\cite{sigasia17FLAME} mesh using PyTorch3D~\cite{ravi2020pytorch3d} to obtain a valid UV mask, from which we sample approximately 130K feature points via bilinear interpolation. These features are then passed through two fully connected layers, followed by separate MLP heads for each Gaussian attribute, decoding the corresponding property values. In practice, we set $\mathrm{N}_{\text{ref}} = 16$ and $\mathrm{N}_{\text{d}} = 8$. We train the model for 150K steps using the Adam~\cite{kingma2017adam} optimizer and a cosine warm-up learning rate scheduler. Our training is conducted on 32 NVIDIA H20 GPUs, taking approximately two weeks to complete.

\noindent \textbf{Datasets.} We train our model on four datasets: VFHQ~\cite{xie2022vfhq}, HDTF~\cite{zhang2021flow}, NeRSemble-v2~\cite{kirschstein2023nersemble} and our synthetic dataset. For all datasets, we obtain pose and expression parameters of FLAME 2023 w/ jaw version and camera parameters using the VHAP~\cite{qian2024vhap} tracker, following the preprocessing protocol in GaussianAvatars~\cite{qian2023gaussianavatars}. To prepare model inputs, we first detect the facial region using the method from GAGAvatar~\cite{chu2024gagavatar}. We then enlarge the bounding box, crop the region of interest, and resize it to $512 \times 512$. We also perform background removal on each input image and randomly replace the background with one of three solid colors: black, white, or gray. For evaluation, we use 50 test clips from VFHQ, together with 25 identity clips split from NeRSemble-v2.

\begin{figure*}[t]
    \centering
    \includegraphics[width=0.95\textwidth]{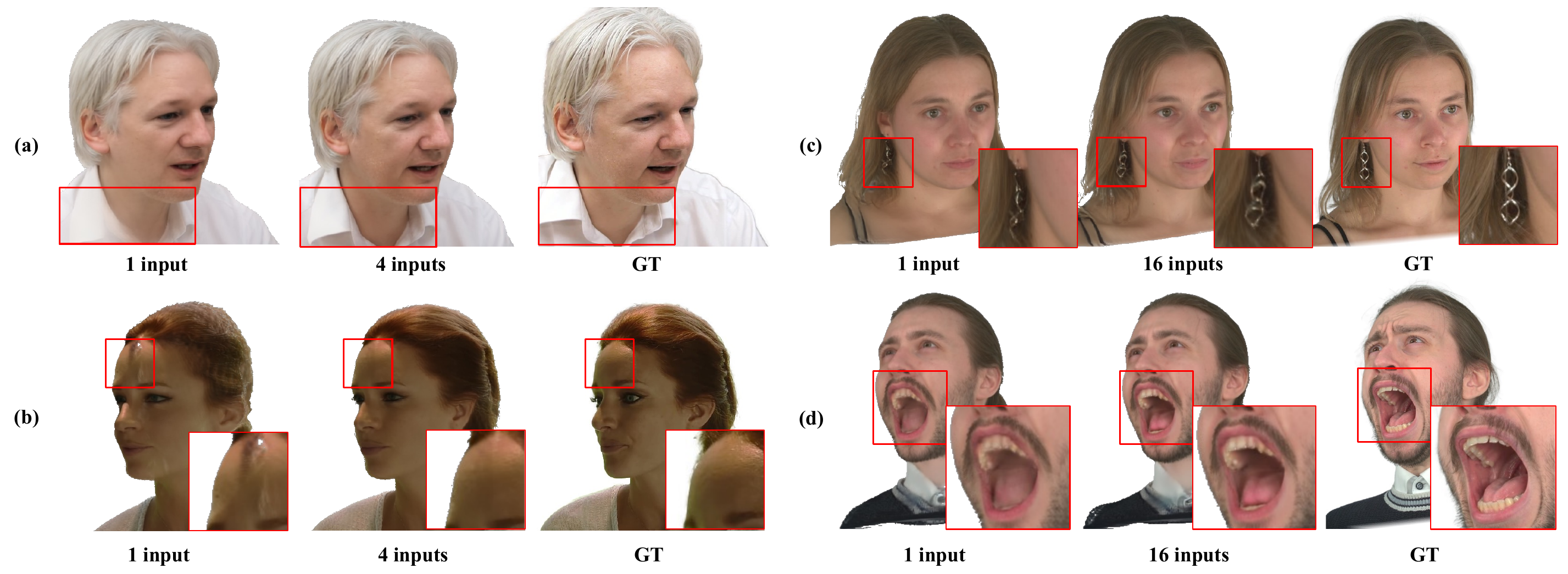}
    \caption{\textbf{Qualitative results of different numbers of input views in VFHQ and NeRSemble-v2 dataset.} 
    }
    \label{fig:view_mono}
\end{figure*}

\begin{figure*}[t]
    \centering
    \includegraphics[width=0.95\textwidth]{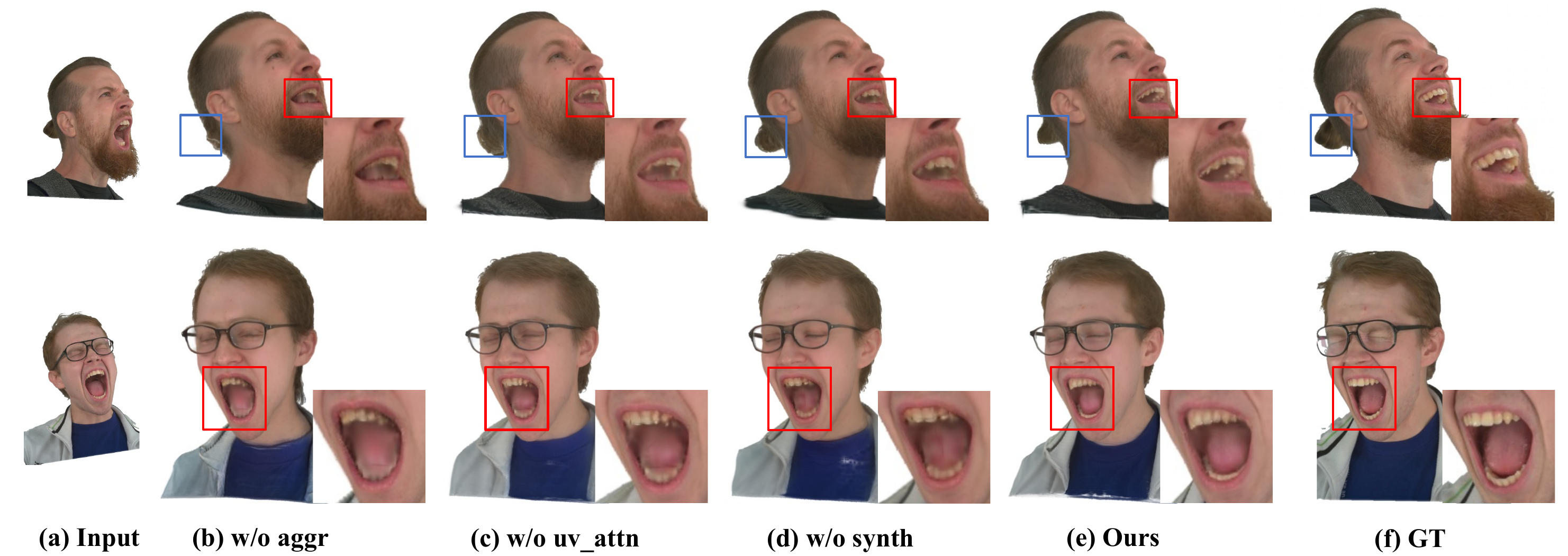}
    \caption{\textbf{Qualitative results for ablation study in the monocular settings in NeRSemble-v2 dataset.} 
    }
    \label{fig:abla_all}
\end{figure*}

\begin{figure}[t!]
    \centering
    \includegraphics[width=0.475\textwidth]{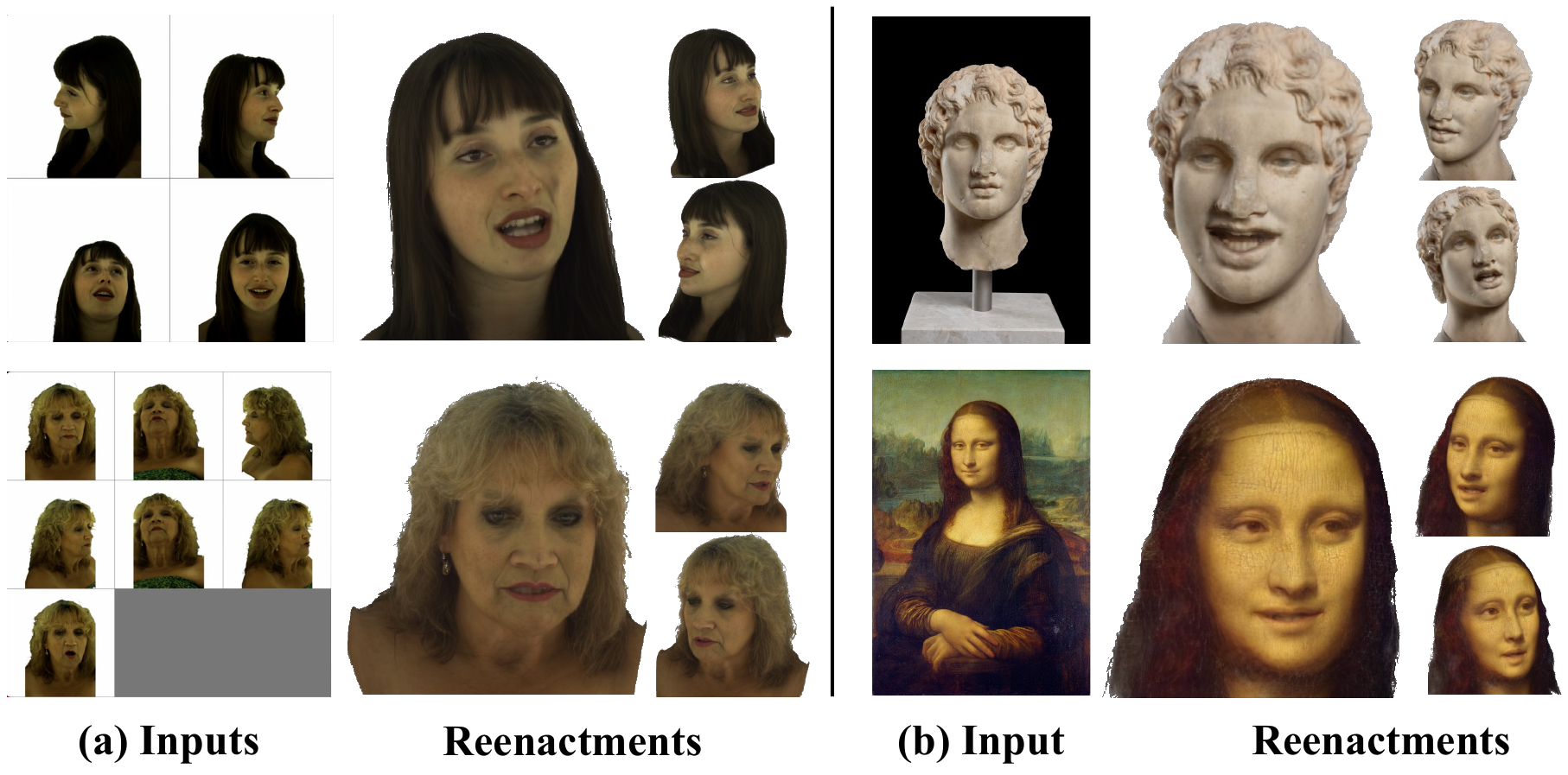}
    \caption{\textbf{Qualitative results for in-the-wild cases.}}
    \label{fig:wild}
\end{figure}

\noindent \textbf{Evaluation Metrics.} We evaluate our model under two input configurations: monocular and multi-view settings. In both cases, we focus on two reenactment scenarios: self reenactment and cross reenactment, and report performance across multiple quantitative metrics. For self reenactment, where ground-truth images are available, we measure image reconstruction quality using PSNR, SSIM, and LPIPS. Identity similarity (CSIM) is computed as the cosine distance between facial feature vectors extracted by ArcFace~\cite{deng2019arcface}. Expression and head pose fidelity are assessed via the Average Expression Distance (AED) and Average Pose Distance (APD), respectively, estimated by the 3DMM-based facial parameter regressor from Deep3DFaceRecon~\cite{deng2019accurate}. Additionally, we measure facial geometry consistency using Average Keypoint Distance (AKD), obtained from a facial landmark detector~\cite{bulat2017far}. For cross reenactment, where ground-truth images are unavailable, we evaluate performance using CSIM, AED, and APD metrics.

\subsection{Main Results}
\label{subsec:res}

\textbf{Baselines.} In the monocular input setting, we compare our approach against state-of-the-art methods, including LAM~\cite{he2025lam}, GAGAvatar~\cite{chu2024gagavatar}, and Portrait4D-v2~\cite{deng2024portrait4dv2}.
In the multi-view input setting, the SOTA baselines are listed as follows: InvertAvatar~\cite{zhao2024invertavatar}, GPAvatar~\cite{chu2024gpavatar}, and DiffusionRig~\cite{ding2023diffusionrig}. Among them, LAM, GAGAvatar, Portrait4D-v2 and GPAvatar are feed-forward methods. However, DiffusionRig requires a fine-tuning phase for each identity, which relies on iterative denoising steps, resulting in slow inference for approximately 30 minutes. Avat3r~\cite{kirschstein2025avat3r} is restricted to a fixed four-view input configuration; due to the absence of an open-source implementation, it is excluded from our baseline comparisons.

\noindent \textbf{Monocular Setting.} 
Our method achieves superior results in Tab.~\ref{tab:comp_metrics1}, especially for self reenactment.
As a conditional generative model, Portrait4D-v2 can not handle extreme expressions in Fig.~\ref{fig:comparison}, because implicit control signal is not efficiently encoded.
Similar to our method, GAGAvatar and LAM utilize canonical Gaussian representations.
When target views are very different from the input source view, GAGAvatar and LAM could degrade the rendering results as shown in Fig.~\ref{fig:comparison}. 
In contrast, our method produces plausible and photo-realistic rendering results on the monocular setting. Further monocular comparison experiments are provided in the supplementary material.

\noindent \textbf{Multi-view Setting.}
It is not trivial to tackle several images in different frames and from different views as inputs.
DiffusionRig and InvertAvatar aggregate latent codes of all input images as conditions to guide a generative model, e.g., a 2D diffusion model or a 3D GAN, to generate the final results.
However, such methods can not efficiently encode the expression from driven images, as shown in Fig.~\ref{fig:comparison}.
Due to the lack of explicit correspondence modeling, GPAvatar and InvertAvatar could even degrade the rendering results when increasing input images.
Thanks to our UV guided modeling, our method outperforms all these baselines in both self and cross reenactments in Tab.~\ref{tab:comp_metrics2}. Further multi-view comparison experiments are provided in the supplementary material.

Thanks to our UV attention branch and self-adaptive fusion strategy, our method is able to aggregate more and more observed information against initial occlusion in Fig.~\ref{fig:view_mono} (a) and to improve 3D consistency in Fig.~\ref{fig:view_mono} (b) as well as rendering details in Fig.~\ref{fig:view_mono} (c, d), when progressively increase number of input images.
Our method also generalizes well to out-of-domain data, including samples from the Ava-256 dataset~\cite{ava256} (Fig.~\ref{fig:wild} (a)) and in-the-wild internet images (Fig.~\ref{fig:wild} (b)). Additional in-the-wild results are provided in the supplementary material.

\begin{table}[t]
\centering
\scalebox{1}{
\begin{tabular}{c|cccc}
\toprule
\textbf{Method} & \textbf{PSNR$\uparrow$} & \textbf{LPIPS$\downarrow$} & \textbf{AED$\downarrow$} & \textbf{AKD$\downarrow$} \\ \hline
w/o synth          & 21.86    & 0.093  & 0.060   & 3.078  \\
w/o uv\_attn     & 22.21    & 0.091  & 0.056   & 3.086  \\
w/o aggr    & 22.39    & 0.088  & 0.059   & 3.120  \\ \hline
Ours              & \textbf{22.61}    & \textbf{0.082}  & \textbf{0.055}  & \textbf{3.037} \\
\bottomrule
\end{tabular}
}
\vspace{-0.1in}
\caption{\textbf{Quantitative results for ablation study on the monocular setting in NeRSemble-v2 datasets for self reenactment.}}
\label{tab:abla_metrics1}
\vspace{-0.15in}
\end{table}

\subsection{Abalation Study}
\label{subsec:abla}

In the following, we study the efficacy of our designed choices for \textsc{UIKA}. The ablations are performed on the monocular setting in the NeRSemble-v2 dataset. Quantitative results are shown in Tab.~\ref{tab:abla_metrics1}.

\noindent \textbf{UV attention branch.}
When removing the UV attention branch from our Transformer network, the learnable UV tokens perform attention only with screen tokens.
Due to the lack of structural information, the ablated version suffers from a significant detail loss in Fig.~\ref{fig:abla_all} (c).

\noindent \textbf{Self-adaptive fusion strategy.} In the ablated version, we do not add the aggregated UV map into our decoding stage. As shown in Fig.~\ref{fig:abla_all} (b), without injection from the observed image, it's hard to yield correct and coherent details.

\noindent \textbf{Importance of our synthetic dataset.} In this version, we trained our model only on VFHQ and NeRSemble-v2 dataset. Comparing to the results in Fig.~\ref{fig:abla_all} (d), our full model, Fig.~\ref{fig:abla_all} (e), preserves view consistency and reconstructs more high-frequency details when using multi-view synthetic data. Further ablation study experiments are provided in the supplementary material.

\section{Conclusion}
\label{sec:con}

In this work, we present \textsc{UIKA}, a feed‑forward framework for animatable Gaussian head avatar modeling from an arbitrary number of pose-free inputs.
By leveraging pixel‑wise facial correspondence estimation, we introduce a UV‑guided avatar modeling pipeline.
We further design a novel UV attention branch to facilitate robust cross-image information matching.
Finally, a self-adaptive fusion strategy is applied to guarantee plausible and complete avatar modeling.
Our method achieves superior results in both monocular and multi-view settings, with a fast run time.

\section*{Acknowledgements}
This work was supported by Ant Group Research Intern Program and the National Natural Science Foundation of China under Grant U25B2046.

{
    \small
    \bibliographystyle{ieeenat_fullname}
    \bibliography{main}

\begin{thebibliography}{111}
\providecommand{\natexlab}[1]{#1}
\providecommand{\url}[1]{\texttt{#1}}
\expandafter\ifx\csname urlstyle\endcsname\relax
  \providecommand{\doi}[1]{doi: #1}\else
  \providecommand{\doi}{doi: \begingroup \urlstyle{rm}\Url}\fi

\bibitem[Buehler et~al.(2024)Buehler, Li, Wood, Helminger, Chen, Shah, Wang, Garbin, Orts-Escolano, Hilliges, et~al.]{buehler2024cafca}
Marcel~C Buehler, Gengyan Li, Erroll Wood, Leonhard Helminger, Xu Chen, Tanmay Shah, Daoye Wang, Stephan Garbin, Sergio Orts-Escolano, Otmar Hilliges, et~al.
\newblock Cafca: High-quality novel view synthesis of expressive faces from casual few-shot captures.
\newblock In \emph{SIGGRAPH Asia 2024 Conference Papers}, pages 1--12, 2024.

\bibitem[Bulat and Tzimiropoulos(2017)]{bulat2017far}
Adrian Bulat and Georgios Tzimiropoulos.
\newblock How far are we from solving the 2d \& 3d face alignment problem? (and a dataset of 230,000 3d facial landmarks).
\newblock In \emph{International Conference on Computer Vision}, 2017.

\bibitem[Burkov et~al.(2020)Burkov, Pasechnik, Grigorev, and Lempitsky]{burkov2020neural}
Egor Burkov, Igor Pasechnik, Artur Grigorev, and Victor Lempitsky.
\newblock Neural head reenactment with latent pose descriptors.
\newblock In \emph{Proceedings of the IEEE/CVF conference on computer vision and pattern recognition}, pages 13786--13795, 2020.

\bibitem[Cai et~al.(2025)Cai, Xiao, Wang, Li, Guo, Fan, Gao, and Zhang]{cai2025hera}
Hongrui Cai, Yuting Xiao, Xuan Wang, Jiafei Li, Yudong Guo, Yanbo Fan, Shenghua Gao, and Juyong Zhang.
\newblock Hera: Hybrid explicit representation for ultra-realistic head avatars.
\newblock In \emph{Proceedings of the Computer Vision and Pattern Recognition Conference}, pages 260--270, 2025.

\bibitem[Chan et~al.(2022)Chan, Lin, Chan, Nagano, Pan, De~Mello, Gallo, Guibas, Tremblay, Khamis, et~al.]{chan2022efficient}
Eric~R Chan, Connor~Z Lin, Matthew~A Chan, Koki Nagano, Boxiao Pan, Shalini De~Mello, Orazio Gallo, Leonidas~J Guibas, Jonathan Tremblay, Sameh Khamis, et~al.
\newblock Efficient geometry-aware 3d generative adversarial networks.
\newblock In \emph{Proceedings of the IEEE/CVF conference on computer vision and pattern recognition}, pages 16123--16133, 2022.

\bibitem[Chen et~al.(2024)Chen, Wang, Li, Xiao, Zhang, Yao, and Liu]{chen2024monogaussianavatar}
Yufan Chen, Lizhen Wang, Qijing Li, Hongjiang Xiao, Shengping Zhang, Hongxun Yao, and Yebin Liu.
\newblock Monogaussianavatar: Monocular gaussian point-based head avatar.
\newblock In \emph{ACM SIGGRAPH 2024 Conference Papers}, pages 1--9, 2024.

\bibitem[Chu and Harada(2024)]{chu2024gagavatar}
Xuangeng Chu and Tatsuya Harada.
\newblock Generalizable and animatable gaussian head avatar.
\newblock In \emph{The Thirty-eighth Annual Conference on Neural Information Processing Systems}, 2024.

\bibitem[Chu et~al.(2024)Chu, Li, Zeng, Yang, Lin, Liu, and Harada]{chu2024gpavatar}
Xuangeng Chu, Yu Li, Ailing Zeng, Tianyu Yang, Lijian Lin, Yunfei Liu, and Tatsuya Harada.
\newblock Gpavatar: Generalizable and precise head avatar from image (s).
\newblock \emph{arXiv preprint arXiv:2401.10215}, 2024.

\bibitem[Deng et~al.(2019{\natexlab{a}})Deng, Guo, Xue, and Zafeiriou]{deng2019arcface}
Jiankang Deng, Jia Guo, Niannan Xue, and Stefanos Zafeiriou.
\newblock Arcface: Additive angular margin loss for deep face recognition.
\newblock In \emph{Proceedings of the IEEE/CVF conference on computer vision and pattern recognition}, pages 4690--4699, 2019{\natexlab{a}}.

\bibitem[Deng et~al.(2019{\natexlab{b}})Deng, Yang, Xu, Chen, Jia, and Tong]{deng2019accurate}
Yu Deng, Jiaolong Yang, Sicheng Xu, Dong Chen, Yunde Jia, and Xin Tong.
\newblock Accurate 3d face reconstruction with weakly-supervised learning: From single image to image set.
\newblock In \emph{Proceedings of the IEEE/CVF conference on computer vision and pattern recognition workshops}, pages 0--0, 2019{\natexlab{b}}.

\bibitem[Deng et~al.(2024{\natexlab{a}})Deng, Wang, Ren, Chen, and Wang]{deng2024portrait4d}
Yu Deng, Duomin Wang, Xiaohang Ren, Xingyu Chen, and Baoyuan Wang.
\newblock Portrait4d: Learning one-shot 4d head avatar synthesis using synthetic data.
\newblock In \emph{Proceedings of the IEEE/CVF Conference on Computer Vision and Pattern Recognition}, pages 7119--7130, 2024{\natexlab{a}}.

\bibitem[Deng et~al.(2024{\natexlab{b}})Deng, Wang, and Wang]{deng2024portrait4dv2}
Yu Deng, Duomin Wang, and Baoyuan Wang.
\newblock Portrait4d-v2: Pseudo multi-view data creates better 4d head synthesizer.
\newblock \emph{arXiv preprint arXiv:2403.13570}, 2024{\natexlab{b}}.

\bibitem[Ding et~al.(2023)Ding, Zhang, Xia, Jebe, Tu, and Zhang]{ding2023diffusionrig}
Zheng Ding, Xuaner Zhang, Zhihao Xia, Lars Jebe, Zhuowen Tu, and Xiuming Zhang.
\newblock Diffusionrig: Learning personalized priors for facial appearance editing.
\newblock In \emph{Proceedings of the IEEE/CVF conference on computer vision and pattern recognition}, pages 12736--12746, 2023.

\bibitem[Esser et~al.(2024)Esser, Kulal, Blattmann, Entezari, M\"{u}ller, Saini, Levi, Lorenz, Sauer, Boesel, Podell, Dockhorn, English, and Rombach]{sd3mmdit}
Patrick Esser, Sumith Kulal, Andreas Blattmann, Rahim Entezari, Jonas M\"{u}ller, Harry Saini, Yam Levi, Dominik Lorenz, Axel Sauer, Frederic Boesel, Dustin Podell, Tim Dockhorn, Zion English, and Robin Rombach.
\newblock Scaling rectified flow transformers for high-resolution image synthesis.
\newblock In \emph{Proceedings of the 41st International Conference on Machine Learning}. JMLR.org, 2024.

\bibitem[Feng et~al.(2021)Feng, Feng, Black, and Bolkart]{feng2021learning}
Yao Feng, Haiwen Feng, Michael~J Black, and Timo Bolkart.
\newblock Learning an animatable detailed 3d face model from in-the-wild images.
\newblock \emph{ACM Transactions on Graphics (ToG)}, 40\penalty0 (4):\penalty0 1--13, 2021.

\bibitem[Gafni et~al.(2021)Gafni, Thies, Zollhofer, and Nie{\ss}ner]{gafni2021nerface}
Guy Gafni, Justus Thies, Michael Zollhofer, and Matthias Nie{\ss}ner.
\newblock Dynamic neural radiance fields for monocular 4d facial avatar reconstruction.
\newblock In \emph{Proceedings of the IEEE/CVF Conference on Computer Vision and Pattern Recognition}, pages 8649--8658, 2021.

\bibitem[Gal et~al.(2022)Gal, Patashnik, Maron, Bermano, Chechik, and Cohen-Or]{gal2022stylegan}
Rinon Gal, Or Patashnik, Haggai Maron, Amit~H Bermano, Gal Chechik, and Daniel Cohen-Or.
\newblock Stylegan-nada: Clip-guided domain adaptation of image generators.
\newblock \emph{ACM Transactions on Graphics (TOG)}, 41\penalty0 (4):\penalty0 1--13, 2022.

\bibitem[Gao et~al.(2025)Gao, Zhou, Liu, Zhou, and Zhang]{Gao2025Learn2Control}
Xuan Gao, Jingtao Zhou, Dongyu Liu, Yuqi Zhou, and Juyong Zhang.
\newblock Constructing diffusion avatar with learnable embeddings.
\newblock In \emph{ACM SIGGRAPH Asia Conference Proceedings}, 2025.

\bibitem[Giebenhain et~al.(2026)Giebenhain, Kirschstein, R{\"u}nz, Agapito, and Nie{\ss}ner]{giebenhain2026pixeldmm}
Simon Giebenhain, Tobias Kirschstein, Martin R{\"u}nz, Lourdes Agapito, and Matthias Nie{\ss}ner.
\newblock Pixel3{DMM}: Versatile screen-space priors for single-image 3d face reconstruction.
\newblock In \emph{The Fourteenth International Conference on Learning Representations}, 2026.

\bibitem[Gong et~al.(2023)Gong, Zhang, Cun, Yin, Fan, Wang, Wu, and Yang]{gong2023toontalker}
Yuan Gong, Yong Zhang, Xiaodong Cun, Fei Yin, Yanbo Fan, Xuan Wang, Baoyuan Wu, and Yujiu Yang.
\newblock Toontalker: Cross-domain face reenactment.
\newblock In \emph{Proceedings of the IEEE/CVF International Conference on Computer Vision}, pages 7690--7700, 2023.

\bibitem[Goodfellow et~al.(2020)Goodfellow, Pouget-Abadie, Mirza, Xu, Warde-Farley, Ozair, Courville, and Bengio]{goodfellow2020generative}
Ian Goodfellow, Jean Pouget-Abadie, Mehdi Mirza, Bing Xu, David Warde-Farley, Sherjil Ozair, Aaron Courville, and Yoshua Bengio.
\newblock Generative adversarial networks.
\newblock \emph{Communications of the ACM}, 63\penalty0 (11):\penalty0 139--144, 2020.

\bibitem[Grassal et~al.(2022)Grassal, Prinzler, Leistner, Rother, Nie{\ss}ner, and Thies]{grassal2022nha}
Philip-William Grassal, Malte Prinzler, Titus Leistner, Carsten Rother, Matthias Nie{\ss}ner, and Justus Thies.
\newblock Neural head avatars from monocular rgb videos.
\newblock In \emph{Proceedings of the IEEE/CVF conference on computer vision and pattern recognition}, pages 18653--18664, 2022.

\bibitem[Guo et~al.(2024)Guo, Zhang, Liu, Zhong, Zhang, Wan, and Zhang]{guo2024liveportrait}
Jianzhu Guo, Dingyun Zhang, Xiaoqiang Liu, Zhizhou Zhong, Yuan Zhang, Pengfei Wan, and Di Zhang.
\newblock Liveportrait: Efficient portrait animation with stitching and retargeting control.
\newblock \emph{arXiv preprint arXiv:2407.03168}, 2024.

\bibitem[He et~al.(2025)He, Gu, Ye, Xu, Zhao, Dong, Yuan, Dong, and Bo]{he2025lam}
Yisheng He, Xiaodong Gu, Xiaodan Ye, Chao Xu, Zhengyi Zhao, Yuan Dong, Weihao Yuan, Zilong Dong, and Liefeng Bo.
\newblock Lam: Large avatar model for one-shot animatable gaussian head.
\newblock In \emph{Proceedings of the Special Interest Group on Computer Graphics and Interactive Techniques Conference Conference Papers}, pages 1--13, 2025.

\bibitem[Hong et~al.(2022{\natexlab{a}})Hong, Zhang, Shen, and Xu]{hong2022depth}
Fa-Ting Hong, Longhao Zhang, Li Shen, and Dan Xu.
\newblock Depth-aware generative adversarial network for talking head video generation.
\newblock In \emph{Proceedings of the IEEE/CVF conference on computer vision and pattern recognition}, pages 3397--3406, 2022{\natexlab{a}}.

\bibitem[Hong et~al.(2022{\natexlab{b}})Hong, Peng, Xiao, Liu, and Zhang]{hong2022headnerf}
Yang Hong, Bo Peng, Haiyao Xiao, Ligang Liu, and Juyong Zhang.
\newblock Headnerf: A real-time nerf-based parametric head model.
\newblock In \emph{Proceedings of the IEEE/CVF Conference on Computer Vision and Pattern Recognition}, pages 20374--20384, 2022{\natexlab{b}}.

\bibitem[Hong et~al.(2023)Hong, Zhang, Gu, Bi, Zhou, Liu, Liu, Sunkavalli, Bui, and Tan]{hong2023lrm}
Yicong Hong, Kai Zhang, Jiuxiang Gu, Sai Bi, Yang Zhou, Difan Liu, Feng Liu, Kalyan Sunkavalli, Trung Bui, and Hao Tan.
\newblock Lrm: Large reconstruction model for single image to 3d.
\newblock \emph{arXiv preprint arXiv:2311.04400}, 2023.

\bibitem[Ji et~al.(2026)Ji, Weiss, Kansy, Naruniec, Cao, Solenthaler, and Bradley]{ji2026fastgha}
Xinya Ji, Sebastian Weiss, Manuel Kansy, Jacek Naruniec, Xun Cao, Barbara Solenthaler, and Derek Bradley.
\newblock Fast{GHA}: Generalized few-shot 3d gaussian head avatars with real-time animation.
\newblock In \emph{The Fourteenth International Conference on Learning Representations}, 2026.

\bibitem[Karras et~al.(2019)Karras, Laine, and Aila]{karras2019style}
Tero Karras, Samuli Laine, and Timo Aila.
\newblock A style-based generator architecture for generative adversarial networks.
\newblock In \emph{Proceedings of the IEEE/CVF conference on computer vision and pattern recognition}, pages 4401--4410, 2019.

\bibitem[Karras et~al.(2021)Karras, Aittala, Laine, H\"ark\"onen, Hellsten, Lehtinen, and Aila]{Karras2021}
Tero Karras, Miika Aittala, Samuli Laine, Erik H\"ark\"onen, Janne Hellsten, Jaakko Lehtinen, and Timo Aila.
\newblock Alias-free generative adversarial networks.
\newblock In \emph{Proc. NeurIPS}, 2021.

\bibitem[Kerbl et~al.(2023)Kerbl, Kopanas, Leimk{\"u}hler, and Drettakis]{kerbl3Dgaussians}
Bernhard Kerbl, Georgios Kopanas, Thomas Leimk{\"u}hler, and George Drettakis.
\newblock 3d gaussian splatting for real-time radiance field rendering.
\newblock \emph{ACM Transactions on Graphics}, 42\penalty0 (4), 2023.

\bibitem[Kingma and Ba(2017)]{kingma2017adam}
Diederik~P. Kingma and Jimmy Ba.
\newblock Adam: A method for stochastic optimization, 2017.

\bibitem[Kirschstein et~al.(2023)Kirschstein, Qian, Giebenhain, Walter, and Nie\ss{}ner]{kirschstein2023nersemble}
Tobias Kirschstein, Shenhan Qian, Simon Giebenhain, Tim Walter, and Matthias Nie\ss{}ner.
\newblock Nersemble: Multi-view radiance field reconstruction of human heads.
\newblock \emph{ACM Trans. Graph.}, 42\penalty0 (4), 2023.

\bibitem[Kirschstein et~al.(2025{\natexlab{a}})Kirschstein, Giebenhain, and Nie{\ss}ner]{kirschstein2025flexavatar}
Tobias Kirschstein, Simon Giebenhain, and Matthias Nie{\ss}ner.
\newblock Flexavatar: Learning complete 3d head avatars with partial supervision.
\newblock \emph{arXiv preprint arXiv:2512.15599}, 2025{\natexlab{a}}.

\bibitem[Kirschstein et~al.(2025{\natexlab{b}})Kirschstein, Romero, Sevastopolsky, Nie{\ss}ner, and Saito]{kirschstein2025avat3r}
Tobias Kirschstein, Javier Romero, Artem Sevastopolsky, Matthias Nie{\ss}ner, and Shunsuke Saito.
\newblock Avat3r: Large animatable gaussian reconstruction model for high-fidelity 3d head avatars.
\newblock \emph{arXiv preprint arXiv:2502.20220}, 2025{\natexlab{b}}.

\bibitem[Lee et~al.(2025)Lee, Kang, Buehler, Kim, Hwang, Hyung, Jang, and Choo]{lee2025surfhead}
Jaeseong Lee, Taewoong Kang, Marcel Buehler, Min-Jung Kim, Sungwon Hwang, Junha Hyung, Hyojin Jang, and Jaegul Choo.
\newblock Surfhead: Affine rig blending for geometrically accurate 2d gaussian surfel head avatars.
\newblock In \emph{The Thirteenth International Conference on Learning Representations}, 2025.

\bibitem[Li et~al.(2024{\natexlab{a}})Li, Chen, Shi, Qiu, An, Chen, and Han]{li2024spherehead}
Heyuan Li, Ce Chen, Tianhao Shi, Yuda Qiu, Sizhe An, Guanying Chen, and Xiaoguang Han.
\newblock Spherehead: Stable 3d full-head synthesis with spherical tri-plane representation, 2024{\natexlab{a}}.

\bibitem[Li et~al.(2017)Li, Bolkart, Black, Li, and Romero]{sigasia17FLAME}
Tianye Li, Timo Bolkart, Michael.~J. Black, Hao Li, and Javier Romero.
\newblock Learning a model of facial shape and expression from {4D} scans.
\newblock \emph{ACM Transactions on Graphics, (Proc. SIGGRAPH Asia)}, 36\penalty0 (6):\penalty0 194:1--194:17, 2017.

\bibitem[Li et~al.(2023)Li, Zhang, Wang, Zhao, Wang, Chen, Zhang, Wang, Bo, and Li]{li2023one}
Weichuang Li, Longhao Zhang, Dong Wang, Bin Zhao, Zhigang Wang, Mulin Chen, Bang Zhang, Zhongjian Wang, Liefeng Bo, and Xuelong Li.
\newblock One-shot high-fidelity talking-head synthesis with deformable neural radiance field.
\newblock In \emph{Proceedings of the IEEE/CVF Conference on Computer Vision and Pattern Recognition}, pages 17969--17978, 2023.

\bibitem[Li et~al.(2024{\natexlab{b}})Li, De~Mello, Liu, Nagano, Iqbal, and Kautz]{li2024generalizable}
Xueting Li, Shalini De~Mello, Sifei Liu, Koki Nagano, Umar Iqbal, and Jan Kautz.
\newblock Generalizable one-shot 3d neural head avatar.
\newblock \emph{Advances in Neural Information Processing Systems}, 36, 2024{\natexlab{b}}.

\bibitem[Liang et~al.(2025)Liang, Ge, Tiwari, Majee, Godaliyadda, Veeraraghavan, and Balakrishnan]{liang2025fastavatar}
Hao Liang, Zhixuan Ge, Ashish Tiwari, Soumendu Majee, GM Godaliyadda, Ashok Veeraraghavan, and Guha Balakrishnan.
\newblock Fastavatar: Instant 3d gaussian splatting for faces from single unconstrained poses.
\newblock \emph{arXiv preprint arXiv:2508.18389}, 2025.

\bibitem[Liao et~al.(2025)Liao, Xu, Li, Li, Zhou, Bai, Xu, Zhang, and Liu]{liao2025hhavatar}
Zhanfeng Liao, Yuelang Xu, Zhe Li, Qijing Li, Boyao Zhou, Ruifeng Bai, Di Xu, Hongwen Zhang, and Yebin Liu.
\newblock Hhavatar: Gaussian head avatar with dynamic hairs.
\newblock \emph{IEEE Transactions on Pattern Analysis and Machine Intelligence}, 2025.

\bibitem[Liu et~al.(2023)Liu, Han, Jin, Qian, Wei, Lin, Wang, Dong, Song, Xu, et~al.]{liu2023human}
Hongyu Liu, Xintong Han, Chengbin Jin, Lihui Qian, Huawei Wei, Zhe Lin, Faqiang Wang, Haoye Dong, Yibing Song, Jia Xu, et~al.
\newblock Human motionformer: Transferring human motions with vision transformers.
\newblock \emph{arXiv preprint arXiv:2302.11306}, 2023.

\bibitem[Liu et~al.(2025)Liu, Wang, Wan, Ma, Chen, Fan, Shen, Song, and Chen]{liu2025avatarartist}
Hongyu Liu, Xuan Wang, Ziyu Wan, Yue Ma, Jingye Chen, Yanbo Fan, Yujun Shen, Yibing Song, and Qifeng Chen.
\newblock Avatarartist: Open-domain 4d avatarization.
\newblock In \emph{Proceedings of the Computer Vision and Pattern Recognition Conference}, pages 10758--10769, 2025.

\bibitem[Ma et~al.(2024{\natexlab{a}})Ma, He, Cun, Wang, Chen, Li, and Chen]{ma2024followyourpose}
Yue Ma, Yingqing He, Xiaodong Cun, Xintao Wang, Siran Chen, Xiu Li, and Qifeng Chen.
\newblock Follow your pose: Pose-guided text-to-video generation using pose-free videos.
\newblock In \emph{Proceedings of the AAAI Conference on Artificial Intelligence}, pages 4117--4125, 2024{\natexlab{a}}.

\bibitem[Ma et~al.(2024{\natexlab{b}})Ma, Liu, Wang, Pan, He, Yuan, Zeng, Cai, Shum, Liu, et~al.]{ma2024followyouremoji}
Yue Ma, Hongyu Liu, Hongfa Wang, Heng Pan, Yingqing He, Junkun Yuan, Ailing Zeng, Chengfei Cai, Heung-Yeung Shum, Wei Liu, et~al.
\newblock Follow-your-emoji: Fine-controllable and expressive freestyle portrait animation.
\newblock \emph{arXiv preprint arXiv:2406.01900}, 2024{\natexlab{b}}.

\bibitem[Ma et~al.(2025)Ma, Yan, Liu, Wang, Pan, He, Yuan, Zeng, Cai, Shum, Li, Liu, linfeng, and Chen]{ma2025followfaster}
Yue Ma, Zexuan Yan, Hongyu Liu, Hongfa Wang, Heng Pan, Yingqing He, Junkun Yuan, Ailing Zeng, Chengfei Cai, Heung-Yeung Shum, Zhifeng Li, Wei Liu, Zhang linfeng, and Qifeng Chen.
\newblock Follow-your-emoji-faster: Towards efficient, fine-controllable, and expressive freestyle portrait animation.
\newblock \emph{International Journal of Computer Vision (IJCV)}, 2025.

\bibitem[Ma et~al.(2023)Ma, Zhu, Qi, Lei, and Zhang]{ma2023otavatar}
Zhiyuan Ma, Xiangyu Zhu, Guo-Jun Qi, Zhen Lei, and Lei Zhang.
\newblock Otavatar: One-shot talking face avatar with controllable tri-plane rendering.
\newblock In \emph{Proceedings of the IEEE/CVF Conference on Computer Vision and Pattern Recognition}, pages 16901--16910, 2023.

\bibitem[Martinez et~al.(2024)Martinez, Kim, Romero, Bagautdinov, Saito, Yu, Anderson, Zollhöfer, Wang, Bai, Li, Wei, Joshi, Borsos, Simon, Saragih, Theodosis, Greene, Josyula, Maeta, Jewett, Venshtain, Heilman, Chen, Fu, Elshaer, Du, Wu, Chen, Kang, Wu, Emad, Longay, Brewer, Shah, Booth, Koska, Haidle, Andromalos, Hsu, Dauer, Selednik, Godisart, Ardisson, Cipperly, Humberston, Farr, Hansen, Guo, Braun, Krenn, Wen, Evans, Fadeeva, Stewart, Schwartz, Gupta, Moon, Guo, Dong, Xu, Shiratori, Prada, Pires, Peng, Buffalini, Trimble, McPhail, Schoeller, and Sheikh]{ava256}
Julieta Martinez, Emily Kim, Javier Romero, Timur Bagautdinov, Shunsuke Saito, Shoou-I Yu, Stuart Anderson, Michael Zollhöfer, Te-Li Wang, Shaojie Bai, Chenghui Li, Shih-En Wei, Rohan Joshi, Wyatt Borsos, Tomas Simon, Jason Saragih, Paul Theodosis, Alexander Greene, Anjani Josyula, Silvio~Mano Maeta, Andrew~I. Jewett, Simon Venshtain, Christopher Heilman, Yueh-Tung Chen, Sidi Fu, Mohamed Ezzeldin~A. Elshaer, Tingfang Du, Longhua Wu, Shen-Chi Chen, Kai Kang, Michael Wu, Youssef Emad, Steven Longay, Ashley Brewer, Hitesh Shah, James Booth, Taylor Koska, Kayla Haidle, Matt Andromalos, Joanna Hsu, Thomas Dauer, Peter Selednik, Tim Godisart, Scott Ardisson, Matthew Cipperly, Ben Humberston, Lon Farr, Bob Hansen, Peihong Guo, Dave Braun, Steven Krenn, He Wen, Lucas Evans, Natalia Fadeeva, Matthew Stewart, Gabriel Schwartz, Divam Gupta, Gyeongsik Moon, Kaiwen Guo, Yuan Dong, Yichen Xu, Takaaki Shiratori, Fabian Prada, Bernardo~R. Pires, Bo Peng, Julia Buffalini, Autumn Trimble, Kevyn McPhail, Melissa Schoeller, and
  Yaser Sheikh.
\newblock {Codec Avatar Studio: Paired Human Captures for Complete, Driveable, and Generalizable Avatars}.
\newblock \emph{NeurIPS Track on Datasets and Benchmarks}, 2024.

\bibitem[Mildenhall et~al.(2020)Mildenhall, Srinivasan, Tancik, Barron, Ramamoorthi, and Ng]{mildenhall2020nerf}
Ben Mildenhall, Pratul~P. Srinivasan, Matthew Tancik, Jonathan~T. Barron, Ravi Ramamoorthi, and Ren Ng.
\newblock Nerf: Representing scenes as neural radiance fields for view synthesis.
\newblock In \emph{ECCV}, 2020.

\bibitem[Mildenhall et~al.(2021)Mildenhall, Srinivasan, Tancik, Barron, Ramamoorthi, and Ng]{mildenhall2021nerf}
Ben Mildenhall, Pratul~P Srinivasan, Matthew Tancik, Jonathan~T Barron, Ravi Ramamoorthi, and Ren Ng.
\newblock Nerf: Representing scenes as neural radiance fields for view synthesis.
\newblock \emph{Communications of the ACM}, 65\penalty0 (1):\penalty0 99--106, 2021.

\bibitem[Mir et~al.(2023)Mir, Alonso, and Mondragón]{mir2023dithead}
Aaron Mir, Eduardo Alonso, and Esther Mondragón.
\newblock Dit-head: High-resolution talking head synthesis using diffusion transformers, 2023.

\bibitem[Oroz et~al.(2025)Oroz, Nießner, and Kirschstein]{oroz2025percheadperceptualheadmodel}
Antonio Oroz, Matthias Nießner, and Tobias Kirschstein.
\newblock Perchead: Perceptual head model for single-image 3d head reconstruction \& editing, 2025.

\bibitem[Pan et~al.(2024)Pan, Zhuo, Piao, Luo, Cheng, Wang, Fan, Liu, Yang, Dai, et~al.]{pan2024renderme360}
Dongwei Pan, Long Zhuo, Jingtan Piao, Huiwen Luo, Wei Cheng, Yuxin Wang, Siming Fan, Shengqi Liu, Lei Yang, Bo Dai, et~al.
\newblock Renderme-360: a large digital asset library and benchmarks towards high-fidelity head avatars.
\newblock \emph{Advances in Neural Information Processing Systems}, 36, 2024.

\bibitem[Paszke et~al.(2019)Paszke, Gross, Massa, Lerer, Bradbury, Chanan, Killeen, Lin, Gimelshein, Antiga, Desmaison, Kopf, Yang, DeVito, Raison, Tejani, Chilamkurthy, Steiner, Fang, Bai, and Chintala]{pytorch}
Adam Paszke, Sam Gross, Francisco Massa, Adam Lerer, James Bradbury, Gregory Chanan, Trevor Killeen, Zeming Lin, Natalia Gimelshein, Luca Antiga, Alban Desmaison, Andreas Kopf, Edward Yang, Zachary DeVito, Martin Raison, Alykhan Tejani, Sasank Chilamkurthy, Benoit Steiner, Lu Fang, Junjie Bai, and Soumith Chintala.
\newblock Pytorch: An imperative style, high-performance deep learning library.
\newblock Curran Associates, Inc., 2019.

\bibitem[Peebles and Xie(2022)]{Peebles2022DiT}
William Peebles and Saining Xie.
\newblock Scalable diffusion models with transformers.
\newblock \emph{arXiv preprint arXiv:2212.09748}, 2022.

\bibitem[Peng et~al.(2025)Peng, Su, Wang, Guo, Li, Long, Lv, Sun, Zhang, and Liu]{peng2025flexavatar}
Cheng Peng, Zhuo Su, Liao Wang, Chen Guo, Zhaohu Li, Chengjiang Long, Zheng Lv, Jingxiang Sun, Chenyangguang Zhang, and Yebin Liu.
\newblock Flexavatar: Flexible large reconstruction model for animatable gaussian head avatars with detailed deformation, 2025.

\bibitem[Qian(2024)]{qian2024vhap}
Shenhan Qian.
\newblock Vhap: Versatile head alignment with adaptive appearance priors, 2024.

\bibitem[Qian et~al.(2024{\natexlab{a}})Qian, Kirschstein, Schoneveld, Davoli, Giebenhain, and Nie\ss{}ner]{qian2023gaussianavatars}
Shenhan Qian, Tobias Kirschstein, Liam Schoneveld, Davide Davoli, Simon Giebenhain, and Matthias Nie\ss{}ner.
\newblock Gaussianavatars: Photorealistic head avatars with rigged 3d gaussians.
\newblock \emph{IEEE Conf. Comput. Vis. Pattern Recog.}, 2024{\natexlab{a}}.

\bibitem[Qian et~al.(2024{\natexlab{b}})Qian, Kirschstein, Schoneveld, Davoli, Giebenhain, and Nie{\ss}ner]{qian2024gaussianavatars}
Shenhan Qian, Tobias Kirschstein, Liam Schoneveld, Davide Davoli, Simon Giebenhain, and Matthias Nie{\ss}ner.
\newblock Gaussianavatars: Photorealistic head avatars with rigged 3d gaussians.
\newblock In \emph{Proceedings of the IEEE/CVF Conference on Computer Vision and Pattern Recognition}, pages 20299--20309, 2024{\natexlab{b}}.

\bibitem[Qiu et~al.(2025{\natexlab{a}})Qiu, Gu, Li, Zuo, Shen, Zhang, Qiu, Yuan, Chen, Dong, and Bo]{Qiu_2025_LHM}
Lingteng Qiu, Xiaodong Gu, Peihao Li, Qi Zuo, Weichao Shen, Junfei Zhang, Kejie Qiu, Weihao Yuan, Guanying Chen, Zilong Dong, and Liefeng Bo.
\newblock Lhm: Large animatable human reconstruction model for single image to 3d in seconds.
\newblock In \emph{Proceedings of the IEEE/CVF International Conference on Computer Vision (ICCV)}, pages 14184--14194, 2025{\natexlab{a}}.

\bibitem[Qiu et~al.(2025{\natexlab{b}})Qiu, Li, Zuo, Gu, Dong, Yuan, Zhu, Han, Chen, and Dong]{qiu2025pf}
Lingteng Qiu, Peihao Li, Qi Zuo, Xiaodong Gu, Yuan Dong, Weihao Yuan, Siyu Zhu, Xiaoguang Han, Guanying Chen, and Zilong Dong.
\newblock Pf-lhm: 3d animatable avatar reconstruction from pose-free articulated human images.
\newblock \emph{arXiv preprint arXiv:2506.13766}, 2025{\natexlab{b}}.

\bibitem[Ranftl et~al.(2020)Ranftl, Lasinger, Hafner, Schindler, and Koltun]{Ranftl2020dpt2}
Ren\'{e} Ranftl, Katrin Lasinger, David Hafner, Konrad Schindler, and Vladlen Koltun.
\newblock Towards robust monocular depth estimation: Mixing datasets for zero-shot cross-dataset transfer.
\newblock \emph{IEEE Transactions on Pattern Analysis and Machine Intelligence (TPAMI)}, 2020.

\bibitem[Ranftl et~al.(2021)Ranftl, Bochkovskiy, and Koltun]{Ranftl2021dpt1}
Ren\'{e} Ranftl, Alexey Bochkovskiy, and Vladlen Koltun.
\newblock Vision transformers for dense prediction.
\newblock \emph{ArXiv preprint}, 2021.

\bibitem[Ravi et~al.(2020)Ravi, Reizenstein, Novotny, Gordon, Lo, Johnson, and Gkioxari]{ravi2020pytorch3d}
Nikhila Ravi, Jeremy Reizenstein, David Novotny, Taylor Gordon, Wan-Yen Lo, Justin Johnson, and Georgia Gkioxari.
\newblock Accelerating 3d deep learning with pytorch3d.
\newblock \emph{arXiv:2007.08501}, 2020.

\bibitem[Shi et~al.(2025)Shi, Li, Wang, Zhu, Cao, and Liu]{shi2025dex}
Yuxiang Shi, Zhe Li, Yanwen Wang, Hao Zhu, Xun Cao, and Ligang Liu.
\newblock Dex-portrait: Disentangled and expressive portrait animation via explicit and latent motion representations.
\newblock \emph{arXiv preprint arXiv:2512.15524}, 2025.

\bibitem[Siarohin et~al.(2019{\natexlab{a}})Siarohin, Lathuili{\`e}re, Tulyakov, Ricci, and Sebe]{siarohin2019animating}
Aliaksandr Siarohin, St{\'e}phane Lathuili{\`e}re, Sergey Tulyakov, Elisa Ricci, and Nicu Sebe.
\newblock Animating arbitrary objects via deep motion transfer.
\newblock In \emph{Proceedings of the IEEE/CVF Conference on Computer Vision and Pattern Recognition}, pages 2377--2386, 2019{\natexlab{a}}.

\bibitem[Siarohin et~al.(2019{\natexlab{b}})Siarohin, Lathuili{\`e}re, Tulyakov, Ricci, and Sebe]{siarohin2019first}
Aliaksandr Siarohin, St{\'e}phane Lathuili{\`e}re, Sergey Tulyakov, Elisa Ricci, and Nicu Sebe.
\newblock First order motion model for image animation.
\newblock \emph{Advances in neural information processing systems}, 32, 2019{\natexlab{b}}.

\bibitem[Siarohin et~al.(2021)Siarohin, Woodford, Ren, Chai, and Tulyakov]{siarohin2021motion}
Aliaksandr Siarohin, Oliver~J Woodford, Jian Ren, Menglei Chai, and Sergey Tulyakov.
\newblock Motion representations for articulated animation.
\newblock In \emph{Proceedings of the IEEE/CVF Conference on Computer Vision and Pattern Recognition}, pages 13653--13662, 2021.

\bibitem[Sim{\'e}oni et~al.(2025)Sim{\'e}oni, Vo, Seitzer, Baldassarre, Oquab, Jose, Khalidov, Szafraniec, Yi, Ramamonjisoa, Massa, Haziza, Wehrstedt, Wang, Darcet, Moutakanni, Sentana, Roberts, Vedaldi, Tolan, Brandt, Couprie, Mairal, J{\'e}gou, Labatut, and Bojanowski]{simeoni2025dinov3}
Oriane Sim{\'e}oni, Huy~V. Vo, Maximilian Seitzer, Federico Baldassarre, Maxime Oquab, Cijo Jose, Vasil Khalidov, Marc Szafraniec, Seungeun Yi, Micha{\"e}l Ramamonjisoa, Francisco Massa, Daniel Haziza, Luca Wehrstedt, Jianyuan Wang, Timoth{\'e}e Darcet, Th{\'e}o Moutakanni, Leonel Sentana, Claire Roberts, Andrea Vedaldi, Jamie Tolan, John Brandt, Camille Couprie, Julien Mairal, Herv{\'e} J{\'e}gou, Patrick Labatut, and Piotr Bojanowski.
\newblock {DINOv3}, 2025.

\bibitem[Stypu{\l}kowski et~al.(2024)Stypu{\l}kowski, Vougioukas, He, Zi{\k{e}}ba, Petridis, and Pantic]{stypulkowski2024diffused}
Micha{\l} Stypu{\l}kowski, Konstantinos Vougioukas, Sen He, Maciej Zi{\k{e}}ba, Stavros Petridis, and Maja Pantic.
\newblock Diffused heads: Diffusion models beat gans on talking-face generation.
\newblock In \emph{Proceedings of the IEEE/CVF Winter Conference on Applications of Computer Vision}, pages 5091--5100, 2024.

\bibitem[Sun et~al.(2023)Sun, Wang, Wang, Li, Zhang, Zhang, and Liu]{sun2023next3d}
Jingxiang Sun, Xuan Wang, Lizhen Wang, Xiaoyu Li, Yong Zhang, Hongwen Zhang, and Yebin Liu.
\newblock Next3d: Generative neural texture rasterization for 3d-aware head avatars.
\newblock In \emph{Proceedings of the IEEE/CVF conference on computer vision and pattern recognition}, pages 20991--21002, 2023.

\bibitem[Taubner et~al.(2025{\natexlab{a}})Taubner, Zhang, Tuli, Bahmani, and Lindell]{taubner2025mvp4d}
Felix Taubner, Ruihang Zhang, Mathieu Tuli, Sherwin Bahmani, and David~B. Lindell.
\newblock Mvp4d: Multi-view portrait video diffusion for animatable 4d avatars.
\newblock New York, NY, USA, 2025{\natexlab{a}}. Association for Computing Machinery.

\bibitem[Taubner et~al.(2025{\natexlab{b}})Taubner, Zhang, Tuli, and Lindell]{taubner2025cap4d}
Felix Taubner, Ruihang Zhang, Mathieu Tuli, and David~B. Lindell.
\newblock {CAP4D}: Creating animatable {4D} portrait avatars with morphable multi-view diffusion models.
\newblock In \emph{Proceedings of the IEEE/CVF Conference on Computer Vision and Pattern Recognition (CVPR)}, pages 5318--5330, 2025{\natexlab{b}}.

\bibitem[Trevithick et~al.(2023)Trevithick, Chan, Stengel, Chan, Liu, Yu, Khamis, Ramamoorthi, and Nagano]{trevithick2023real}
Alex Trevithick, Matthew Chan, Michael Stengel, Eric Chan, Chao Liu, Zhiding Yu, Sameh Khamis, Ravi Ramamoorthi, and Koki Nagano.
\newblock Real-time radiance fields for single-image portrait view synthesis.
\newblock 2023.

\bibitem[Wang et~al.(2023)Wang, Deng, Yin, Shum, and Wang]{wang2023progressive}
Duomin Wang, Yu Deng, Zixin Yin, Heung-Yeung Shum, and Baoyuan Wang.
\newblock Progressive disentangled representation learning for fine-grained controllable talking head synthesis.
\newblock In \emph{Proceedings of the IEEE/CVF Conference on Computer Vision and Pattern Recognition}, pages 17979--17989, 2023.

\bibitem[Wang et~al.(2025{\natexlab{a}})Wang, Chen, Karaev, Vedaldi, Rupprecht, and Novotny]{wang2025vggt}
Jianyuan Wang, Minghao Chen, Nikita Karaev, Andrea Vedaldi, Christian Rupprecht, and David Novotny.
\newblock Vggt: Visual geometry grounded transformer.
\newblock In \emph{Proceedings of the IEEE/CVF Conference on Computer Vision and Pattern Recognition}, 2025{\natexlab{a}}.

\bibitem[Wang et~al.(2025{\natexlab{b}})Wang, Zhuang, Wang, Cao, Guo, Zuo, and Zhu]{Sketch2PoseNet2025wang}
Li Wang, Yiyu Zhuang, Yanwen Wang, Xun Cao, Chuan Guo, Xinxin Zuo, and Hao Zhu.
\newblock Sketch2posenet: Efficient and generalized sketch to 3d human pose prediction.
\newblock In \emph{Proceedings of the SIGGRAPH Asia 2025 Conference Papers}, New York, NY, USA, 2025{\natexlab{b}}. Association for Computing Machinery.

\bibitem[Wang et~al.(2024)Wang, Tan, Bi, Xu, Luan, Sunkavalli, Wang, Xu, and Zhang]{wang2024pf}
Peng Wang, Hao Tan, Sai Bi, Yinghao Xu, Fujun Luan, Kalyan Sunkavalli, Wenping Wang, Zexiang Xu, and Kai Zhang.
\newblock Pf-lrm: Pose-free large reconstruction model for joint pose and shape prediction.
\newblock In \emph{ICLR}, 2024.

\bibitem[Wang et~al.(2021)Wang, Li, Zhang, and Shan]{gfpgan}
Xintao Wang, Yu Li, Honglun Zhang, and Ying Shan.
\newblock Towards real-world blind face restoration with generative facial prior.
\newblock In \emph{IEEE Conf. Comput. Vis. Pattern Recog.}, 2021.

\bibitem[Wang et~al.(2025{\natexlab{c}})Wang, Wang, Yi, Fan, Hu, Zhu, and Ma]{wang20253d}
Yating Wang, Xuan Wang, Ran Yi, Yanbo Fan, Jichen Hu, Jingcheng Zhu, and Lizhuang Ma.
\newblock 3d gaussian head avatars with expressive dynamic appearances by compact tensorial representations.
\newblock In \emph{Proceedings of the Computer Vision and Pattern Recognition Conference}, pages 21117--21126, 2025{\natexlab{c}}.

\bibitem[Wang et~al.(2025{\natexlab{d}})Wang, Zhuang, Zhang, Wang, Zeng, Cao, Zuo, and Zhu]{wang2025tera}
Yanwen Wang, Yiyu Zhuang, Jiawei Zhang, Li Wang, Yifei Zeng, Xun Cao, Xinxin Zuo, and Hao Zhu.
\newblock Tera: Rethinking text-guided realistic 3d avatar generation.
\newblock In \emph{Proceedings of the IEEE/CVF International Conference on Computer Vision}, pages 10686--10697, 2025{\natexlab{d}}.

\bibitem[Wei et~al.(2024)Wei, Yang, and Wang]{wei2024aniportrait}
Huawei Wei, Zejun Yang, and Zhisheng Wang.
\newblock Aniportrait: Audio-driven synthesis of photorealistic portrait animations.
\newblock \emph{arXiv:2403.17694}, 2024.

\bibitem[Wu et~al.(2026)Wu, Chen, Wu, Li, Lu, and Feng]{wu2026fastavatar}
Yue Wu, Xuanhong Chen, Yufan Wu, Wen Li, Yuxi Lu, and Kairui Feng.
\newblock Fastavatar: Towards unified and fast 3d avatar reconstruction with large gaussian reconstruction transformers.
\newblock In \emph{The Fourteenth International Conference on Learning Representations}, 2026.

\bibitem[Xiang et~al.(2024)Xiang, Gao, Guo, and Zhang]{xiang2024flashavatar}
Jun Xiang, Xuan Gao, Yudong Guo, and Juyong Zhang.
\newblock Flashavatar: High-fidelity head avatar with efficient gaussian embedding.
\newblock In \emph{Proceedings of the IEEE/CVF Conference on Computer Vision and Pattern Recognition}, pages 1802--1812, 2024.

\bibitem[Xie et~al.(2022)Xie, Wang, Zhang, Dong, and Shan]{xie2022vfhq}
Liangbin Xie, Xintao Wang, Honglun Zhang, Chao Dong, and Ying Shan.
\newblock Vfhq: A high-quality dataset and benchmark for video face super-resolution.
\newblock In \emph{The IEEE Conference on Computer Vision and Pattern Recognition Workshops (CVPRW)}, 2022.

\bibitem[Xie et~al.(2024)Xie, Xu, Song, Wang, Shi, and Luo]{xie2024x}
You Xie, Hongyi Xu, Guoxian Song, Chao Wang, Yichun Shi, and Linjie Luo.
\newblock X-portrait: Expressive portrait animation with hierarchical motion attention.
\newblock In \emph{ACM SIGGRAPH 2024 Conference Papers}, pages 1--11, 2024.

\bibitem[Xu et~al.(2024{\natexlab{a}})Xu, Chen, Guo, Yang, Li, Zang, Zhang, Tong, and Guo]{xu2024vasa}
Sicheng Xu, Guojun Chen, Yu-Xiao Guo, Jiaolong Yang, Chong Li, Zhenyu Zang, Yizhong Zhang, Xin Tong, and Baining Guo.
\newblock {VASA}-1: Lifelike audio-driven talking faces generated in real time.
\newblock In \emph{The Thirty-eighth Annual Conference on Neural Information Processing Systems}, 2024{\natexlab{a}}.

\bibitem[Xu et~al.(2023)Xu, Wang, Zhao, Zhang, and Liu]{xu2023avatarmav}
Yuelang Xu, Lizhen Wang, Xiaochen Zhao, Hongwen Zhang, and Yebin Liu.
\newblock Avatarmav: Fast 3d head avatar reconstruction using motion-aware neural voxels.
\newblock In \emph{ACM SIGGRAPH 2023 Conference Proceedings}, pages 1--10, 2023.

\bibitem[Xu et~al.(2024{\natexlab{b}})Xu, Chen, Li, Zhang, Wang, Zheng, and Liu]{Xu_2024_CVPR}
Yuelang Xu, Benwang Chen, Zhe Li, Hongwen Zhang, Lizhen Wang, Zerong Zheng, and Yebin Liu.
\newblock Gaussian head avatar: Ultra high-fidelity head avatar via dynamic gaussians.
\newblock In \emph{Proceedings of the IEEE/CVF Conference on Computer Vision and Pattern Recognition (CVPR)}, pages 1931--1941, 2024{\natexlab{b}}.

\bibitem[Xu et~al.(2024{\natexlab{c}})Xu, Chen, Li, Zhang, Wang, Zheng, and Liu]{xu2023gaussianheadavatar}
Yuelang Xu, Benwang Chen, Zhe Li, Hongwen Zhang, Lizhen Wang, Zerong Zheng, and Yebin Liu.
\newblock Gaussian head avatar: Ultra high-fidelity head avatar via dynamic gaussians.
\newblock In \emph{IEEE Conf. Comput. Vis. Pattern Recog.}, 2024{\natexlab{c}}.

\bibitem[Xu et~al.(2024{\natexlab{d}})Xu, Shi, Yifan, Chen, Yang, Peng, Shen, and Wetzstein]{xu2024grm}
Yinghao Xu, Zifan Shi, Wang Yifan, Hansheng Chen, Ceyuan Yang, Sida Peng, Yujun Shen, and Gordon Wetzstein.
\newblock Grm: Large gaussian reconstruction model for efficient 3d reconstruction and generation.
\newblock \emph{arXiv preprint arXiv:2403.14621}, 2024{\natexlab{d}}.

\bibitem[Xu et~al.(2025)Xu, Wang, Zheng, Su, and Liu]{xu2025gphm}
Yuelang Xu, Lizhen Wang, Zerong Zheng, Zhaoqi Su, and Yebin Liu.
\newblock 3d gaussian parametric head model.
\newblock In \emph{European Conference on Computer Vision}, pages 129--147. Springer, 2025.

\bibitem[Yang et~al.(2020)Yang, Zhu, Wang, Huang, Shen, Yang, and Cao]{yang2020facescape}
Haotian Yang, Hao Zhu, Yanru Wang, Mingkai Huang, Qiu Shen, Ruigang Yang, and Xun Cao.
\newblock Facescape: a large-scale high quality 3d face dataset and detailed riggable 3d face prediction.
\newblock In \emph{Proceedings of the ieee/cvf conference on computer vision and pattern recognition}, pages 601--610, 2020.

\bibitem[Ye et~al.(2024)Ye, Zhong, Ren, Yang, Li, Huang, Jiang, He, Huang, Liu, et~al.]{ye2024real3d}
Zhenhui Ye, Tianyun Zhong, Yi Ren, Jiaqi Yang, Weichuang Li, Jiawei Huang, Ziyue Jiang, Jinzheng He, Rongjie Huang, Jinglin Liu, et~al.
\newblock Real3d-portrait: One-shot realistic 3d talking portrait synthesis.
\newblock \emph{arXiv preprint arXiv:2401.08503}, 2024.

\bibitem[Yu et~al.(2025)Yu, Zhu, and Cao]{yu2025realityavatar}
Houteng Yu, Hao Zhu, and Xun Cao.
\newblock { RealityAvatar: Comprehensive Head Avatar Generation with 360° Rendering }.
\newblock In \emph{2025 IEEE International Conference on Multimedia and Expo (ICME)}, pages 1--6, Los Alamitos, CA, USA, 2025. IEEE Computer Society.

\bibitem[Yu et~al.(2023{\natexlab{a}})Yu, Fan, Zhang, Wang, Yin, Bai, Cao, Shan, Wu, Sun, et~al.]{yu2023nofa}
Wangbo Yu, Yanbo Fan, Yong Zhang, Xuan Wang, Fei Yin, Yunpeng Bai, Yan-Pei Cao, Ying Shan, Yang Wu, Zhongqian Sun, et~al.
\newblock Nofa: Nerf-based one-shot facial avatar reconstruction.
\newblock In \emph{ACM SIGGRAPH 2023 Conference Proceedings}, pages 1--12, 2023{\natexlab{a}}.

\bibitem[Yu et~al.(2023{\natexlab{b}})Yu, Yin, Zhou, Wang, Wong, and Wang]{yu2023thpad}
Zhentao Yu, Zixin Yin, Deyu Zhou, Duomin Wang, Finn Wong, and Baoyuan Wang.
\newblock Talking head generation with probabilistic audio-to-visual diffusion priors.
\newblock In \emph{International Conference on Computer Vision (ICCV)}, 2023{\natexlab{b}}.

\bibitem[Yu et~al.(2024)Yu, Bai, Meka, Tan, Xu, Pandey, Fanello, Park, and Zhang]{yu2024one2avatar}
Zhixuan Yu, Ziqian Bai, Abhimitra Meka, Feitong Tan, Qiangeng Xu, Rohit Pandey, Sean Fanello, Hyun~Soo Park, and Yinda Zhang.
\newblock One2avatar: Generative implicit head avatar for few-shot user adaptation.
\newblock \emph{arXiv preprint arXiv:2402.11909}, 2024.

\bibitem[Zakharov et~al.(2019)Zakharov, Shysheya, Burkov, and Lempitsky]{zakharov2019few}
Egor Zakharov, Aliaksandra Shysheya, Egor Burkov, and Victor Lempitsky.
\newblock Few-shot adversarial learning of realistic neural talking head models.
\newblock In \emph{Proceedings of the IEEE/CVF international conference on computer vision}, pages 9459--9468, 2019.

\bibitem[Zhang et~al.(2023)Zhang, Qi, Zhang, Zhang, Wu, Chen, Chen, Wang, and Wen]{zhang2023metaportrait}
Bowen Zhang, Chenyang Qi, Pan Zhang, Bo Zhang, HsiangTao Wu, Dong Chen, Qifeng Chen, Yong Wang, and Fang Wen.
\newblock Metaportrait: Identity-preserving talking head generation with fast personalized adaptation.
\newblock In \emph{Proceedings of the IEEE/CVF Conference on Computer Vision and Pattern Recognition}, pages 22096--22105, 2023.

\bibitem[Zhang et~al.(2025{\natexlab{a}})Zhang, Chu, Li, Zang, Li, Li, Cao, Zhu, and Lu]{zhang2025bringingportrait3dpresence}
Jiawei Zhang, Lei Chu, Jiahao Li, Zhenyu Zang, Chong Li, Xiao Li, Xun Cao, Hao Zhu, and Yan Lu.
\newblock Bringing your portrait to 3d presence.
\newblock \emph{arXiv preprint arXiv:2511.22553}, 2025{\natexlab{a}}.

\bibitem[Zhang et~al.(2025{\natexlab{b}})Zhang, Wu, Liang, Gong, Hu, Yao, Cao, and Zhu]{zhang2025fate}
Jiawei Zhang, Zijian Wu, Zhiyang Liang, Yicheng Gong, Dongfang Hu, Yao Yao, Xun Cao, and Hao Zhu.
\newblock Fate: Full-head gaussian avatar with textural editing from monocular video.
\newblock In \emph{Proceedings of the Computer Vision and Pattern Recognition Conference}, pages 5535--5545, 2025{\natexlab{b}}.

\bibitem[Zhang et~al.(2021)Zhang, Li, Ding, and Fan]{zhang2021flow}
Zhimeng Zhang, Lincheng Li, Yu Ding, and Changjie Fan.
\newblock Flow-guided one-shot talking face generation with a high-resolution audio-visual dataset.
\newblock In \emph{Proceedings of the IEEE/CVF Conference on Computer Vision and Pattern Recognition}, pages 3661--3670, 2021.

\bibitem[Zhao and Xu(2026)]{zhao2026generalizable}
Shuling Zhao and Dan Xu.
\newblock Generalizable and animatable 3d full-head gaussian avatar from a single image.
\newblock \emph{arXiv preprint arXiv:2601.12770}, 2026.

\bibitem[Zhao et~al.(2024)Zhao, Sun, Wang, Suo, and Liu]{zhao2024invertavatar}
Xiaochen Zhao, Jingxiang Sun, Lizhen Wang, Jinli Suo, and Yebin Liu.
\newblock Invertavatar: Incremental gan inversion for generalized head avatars.
\newblock In \emph{ACM SIGGRAPH 2024 Conference Papers}, pages 1--10, 2024.

\bibitem[Zhao et~al.(2025)Zhao, Xu, Song, Xie, Zhang, Li, Luo, Suo, and Liu]{zhao2025xnemoexpressiveneuralmotion}
Xiaochen Zhao, Hongyi Xu, Guoxian Song, You Xie, Chenxu Zhang, Xiu Li, Linjie Luo, Jinli Suo, and Yebin Liu.
\newblock X-nemo: Expressive neural motion reenactment via disentangled latent attention, 2025.

\bibitem[Zheng et~al.(2024)Zheng, Wen, Li, Zhang, Su, Chang, Zhao, Lv, Zhang, Zhang, et~al.]{zheng2024headgap}
Xiaozheng Zheng, Chao Wen, Zhaohu Li, Weiyi Zhang, Zhuo Su, Xu Chang, Yang Zhao, Zheng Lv, Xiaoyuan Zhang, Yongjie Zhang, et~al.
\newblock Headgap: Few-shot 3d head avatar via generalizable gaussian priors.
\newblock \emph{arXiv preprint arXiv:2408.06019}, 2024.

\bibitem[Zhuang et~al.(2022)Zhuang, Zhu, Sun, and Cao]{zhuang2022mofanerf}
Yiyu Zhuang, Hao Zhu, Xusen Sun, and Xun Cao.
\newblock Mofanerf: Morphable facial neural radiance field.
\newblock In \emph{European conference on computer vision}, pages 268--285. Springer, 2022.

\bibitem[Zhuang et~al.(2024)Zhuang, Lv, Wen, Shuai, Zeng, Zhu, Chen, Yang, Cao, and Liu]{zhuang2024idolinstant}
Yiyu Zhuang, Jiaxi Lv, Hao Wen, Qing Shuai, Ailing Zeng, Hao Zhu, Shifeng Chen, Yujiu Yang, Xun Cao, and Wei Liu.
\newblock Idol: Instant photorealistic 3d human creation from a single image.
\newblock \emph{arXiv preprint arXiv:2412.14963}, 2024.

\bibitem[Zielonka et~al.(2023)Zielonka, Bolkart, and Thies]{zielonka2023insta}
Wojciech Zielonka, Timo Bolkart, and Justus Thies.
\newblock Instant volumetric head avatars.
\newblock In \emph{Proceedings of the IEEE/CVF conference on computer vision and pattern recognition}, pages 4574--4584, 2023.

\end{thebibliography}
}

\clearpage
\maketitlesupplementary

\renewcommand\thesection{\Alph{section}}
\setcounter{section}{0}
\renewcommand\thefigure{S\arabic{figure}}
\renewcommand\thetable{S\arabic{table}}
\setcounter{figure}{0}
\setcounter{table}{0}


In this supplementary material, we first provide additional implementation details for our model, along with visualizations of components (Sec.~\ref{sec:implement_suppl}). We then present and evaluate our synthetic dataset (Sec.~\ref{sec:synth_suppl}). Next, we report additional comparative experiments and a user study, covering both self and cross reenactment on monocular and multi‑view settings (Sec.~\ref{sec:comp_suppl}). We also include extended ablation studies, examining the impact of training data size as well as ablations of our method itself (Sec.~\ref{sec:abla_suppl}). We then provide more in‑the‑wild cases and applications, e.g., text‑to‑head‑avatar generation (Sec.~\ref{sec:app_suppl}). Finally, we discuss the limitations of our method (Sec.~\ref{sec:limit}) and its associated ethical implications (Sec.~\ref{sec:ethics}). Additional dynamic results are provided in our supplementary video.

\section{Additional Implementation Details}
\label{sec:implement_suppl}

\subsection{Facial Correspondence Estimator}
\label{subsec:uv_pred_suppl}

\begin{figure*}[htbp!]
    \centering
    \includegraphics[width=1.0\textwidth]{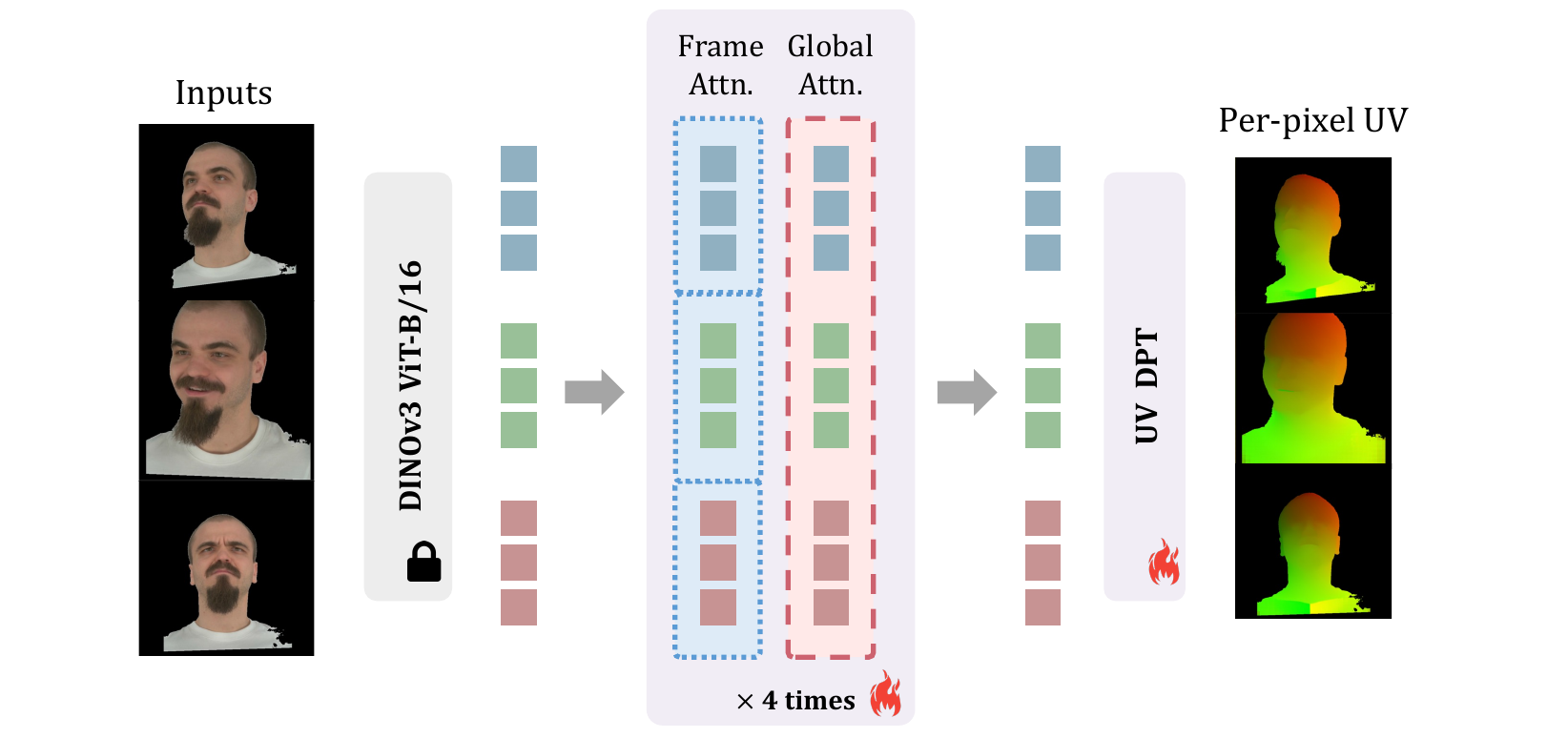}
    \caption{\textbf{Architecture of our facial correspondence estimator network.}}
    \label{fig:uv_pred_net}
\end{figure*}

As illustrated in Fig.~\ref{fig:uv_pred_net}, our facial correspondence estimator network consists of three main components: a frozen feature extractor, a trainable alternating attention module, and a trainable UV decoding head. We first extract patch-wise features using DINOv3 ViT-B/16~\cite{simeoni2025dinov3}, which serves as a powerful pre-trained visual backbone. The extracted features are processed through \textit{four} alternating attention layers. Following VGGT~\cite{wang2025vggt}, \textit{Frame Attention} first captures intra-frame spatial relationships within each individual image by computing \textit{self attention} across patches of the same frame. \textit{Global Attention} then establishes inter-frame correspondences by attending to tokens across all input frames simultaneously. This hierarchical attention design enables the network to jointly reason about local facial structures and global multi-view consistency. On top of it, we initialize a trainable DPT~\cite{Ranftl2021dpt1, Ranftl2020dpt2} head to predict two‑channel UV coordinates within the range $[0, 1]$. The predicted UV coordinates map is further multiplied by the input image mask to extract the valid foreground region of the human head.

\subsection{UV coordinates map}
\label{subsec:uv_pred_res}

\begin{figure}[t!]
    \centering
    \includegraphics[width=0.5\textwidth]{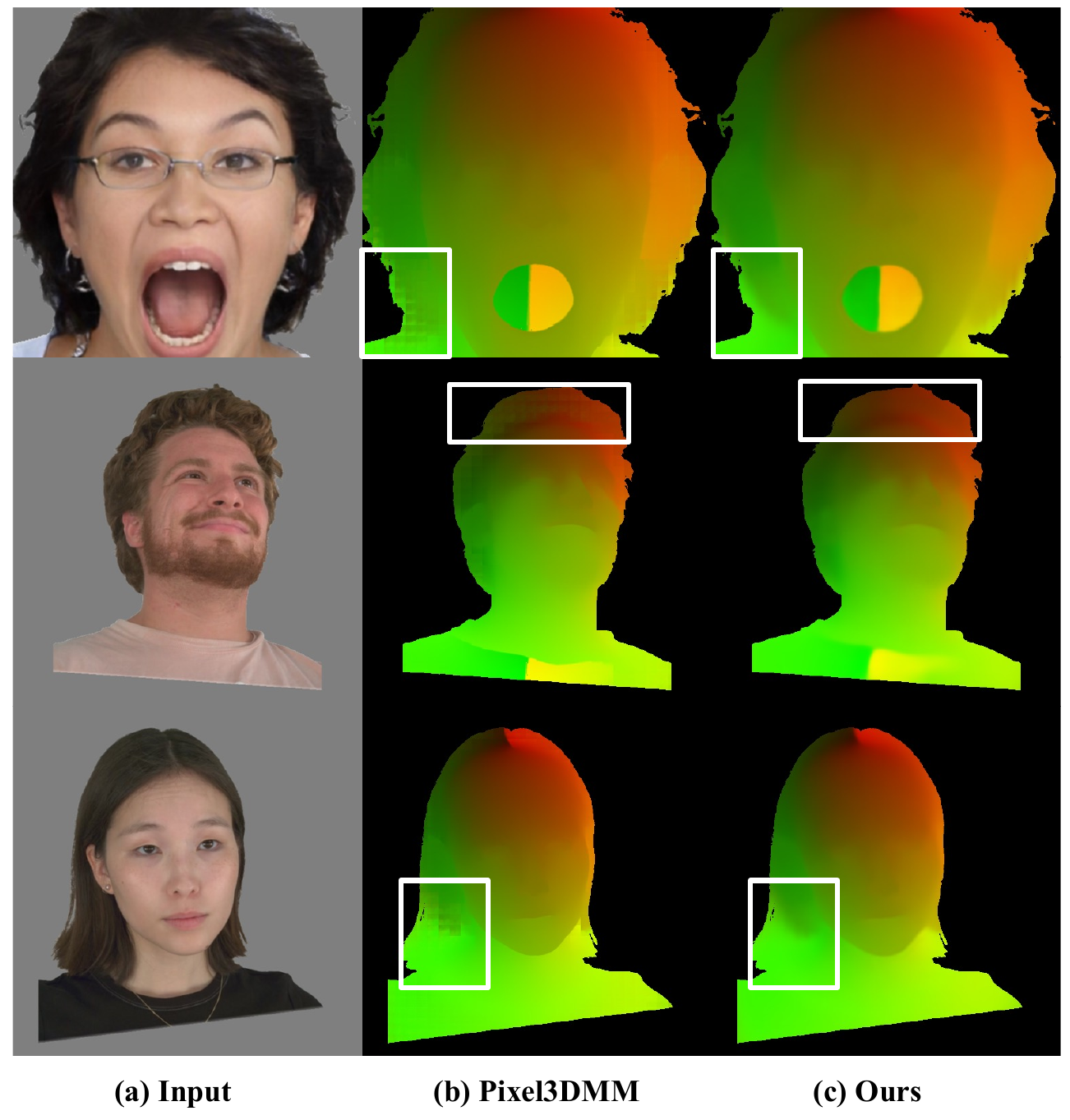}
    \caption{\textbf{Visualization and Comparison of UV coordinates map.}}
    \label{fig:uv_pred_res}
\end{figure}

We visualize the predicted UV coordinates map and compare them with Pixel3DMM~\cite{giebenhain2026pixeldmm}. As shown in Fig.~\ref{fig:uv_pred_res}, our approach produces significantly smoother results in the boundary regions, particularly around the hair. This smoothness is crucial for our subsequent reprojecting of screen-space color back into UV space, enabling more coherent and reliable reprojection results.

\subsection{Hyperparameters}
\label{subsec:hyper}

\begin{table}[t]
\centering
\scalebox{0.87}{
\begin{tabular}{c|c|c}
\toprule
 & \textbf{Hyperparameter} & \textbf{Value} \\ \hline

\multirow{2}{1.8cm}{\textbf{Input \&} \\ \textbf{Output}} & Input image resolution & 512 $\times$ 512 \\
 & Train render resolution & 512 $\times$ 512 \\ \hline

\multirow{4}{1.8cm}{\textbf{Feature} \\ \textbf{Extractor}} & DINOv3 version & vitl16 \\
 & DINOv3 patch size & 16 $\times$ 16 \\
 & DINOv3 feature size & $\mathcal{N}$ $\times$ 1024 $\times$ 1024 \\
 & DINOv3 intermediate layer & 4, 11, 17, 23 \\ \hline

\multirow{4}{1.8cm}{\textbf{MultiModal} \\ \textbf{Transformer}} & Hidden dimension & 1024 \\
 & Head numbers & 16 \\
 & Self attention layers & 12 \\
 & Learnable UV token size & 96 $\times$ 96 $\times$ 1024 \\ \hline

\multirow{6}{1.8cm}{\textbf{UV} \\ \textbf{Gaussian} \\ \textbf{Decoder}} & Gaussian attribute map size & 384 $\times$ 384 \\
 & Aggregated UV map size & 384 $\times$ 384 \\
 & UV DPT inner dimension & 256 \\
 & MLP inner dimension & 512 \\
 & MLP layers & 3 \\
 & MLP activation & SiLU \\ \hline

\multirow{4}{1.8cm}{\textbf{Gaussian} \\ \textbf{Settings}} & Offset max range & 0.2 \\
 & Scaling clip range & 0.01 \\
 & Init scaling & exp(-5.0) \\
 & Init density & 0.1 \\ 

\bottomrule
\end{tabular}
}
\caption{\textbf{Hyperparameters used in our method.} $\mathcal{N}$ represents the number of input views.}
\label{tab:hyper}
\end{table}

In Tab. \ref{tab:hyper}, we provide additional detailed hyperparameters used in our model configuration.

\subsection{Latency analysis.}
\label{subsec:latency}

\begin{table}[t]
\centering
\scalebox{0.9}{
\begin{tabular}{c|ccccccc}
\toprule
\textbf{Input number} & \textbf{1} & \textbf{2} & \textbf{4} & \textbf{8} & \textbf{16} & \textbf{32}  \\ \hline
\textbf{V-D Latency (s)}     & 1.96 & 2.51 & 3.57 & 6.02 & 12.8 & 32.9 \\
\bottomrule
\end{tabular}
}
\caption{\textbf{Latency analysis for view-dependent (V-D) modules.} We show running time for different number of input images.}
\label{tab:latency}
\end{table}

In general, one pass inference consists of view-dependent (V-D) and view-independent (V-I) components. As shown in Tab.~\ref{tab:latency}, the latency of V-D components (UV prediction, Transformer, and decoder) scales in $O\left( N^2\right)$ with the number of input images inherent to the self-attention mechanism. Once the canonical Gaussian avatar is obtained, the V-I module takes 3ms for LBS and 2ms for rendering. The total latency is in second level.

\begin{figure}[t!]
    \centering
    \includegraphics[width=0.5\textwidth]{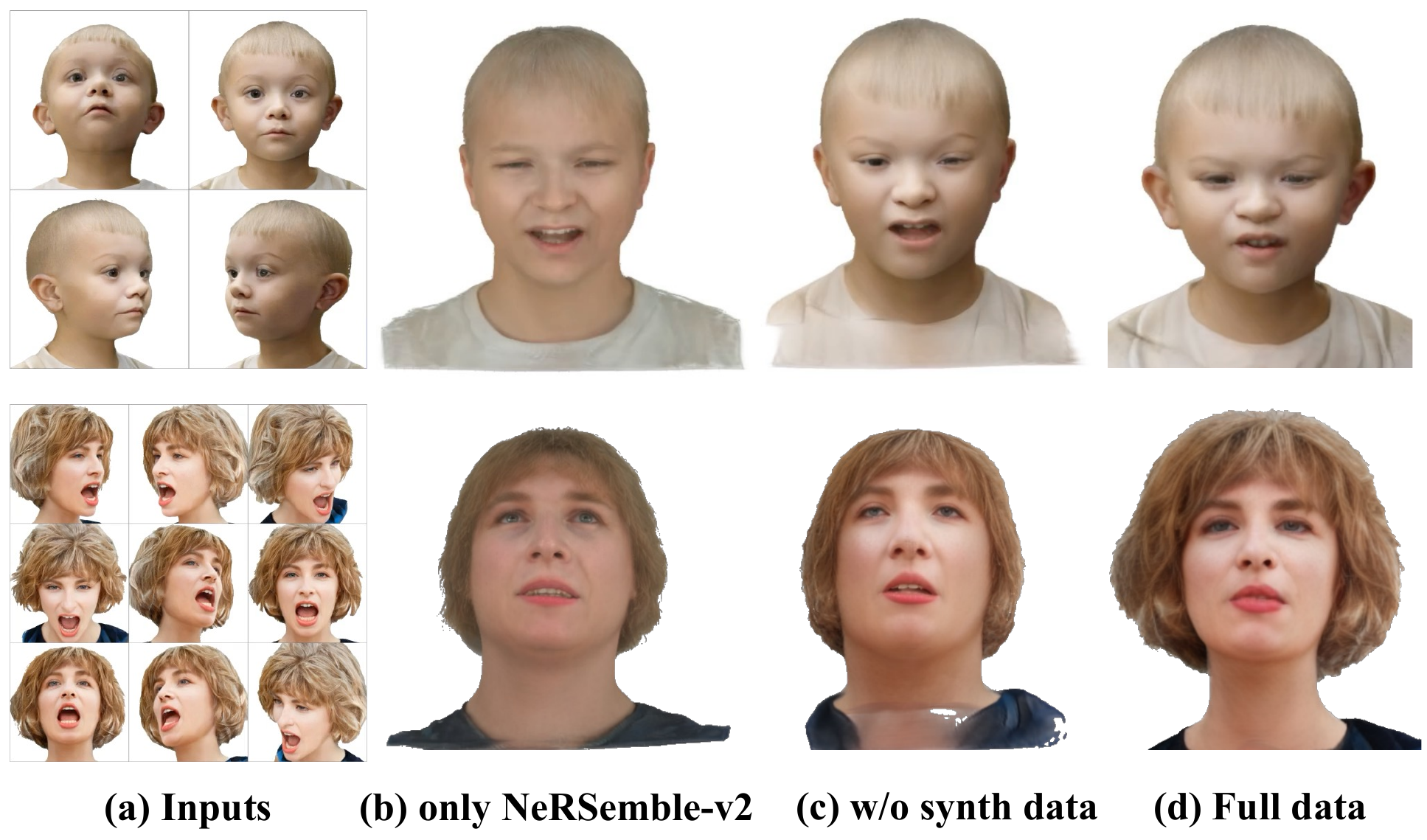}
    \caption{\textbf{Ablation study on identity robustness across different training data configurations.}}
    \label{fig:abla_data_suppl}
\end{figure}

\begin{figure}[t!]
    \centering
    \includegraphics[width=0.47\textwidth]{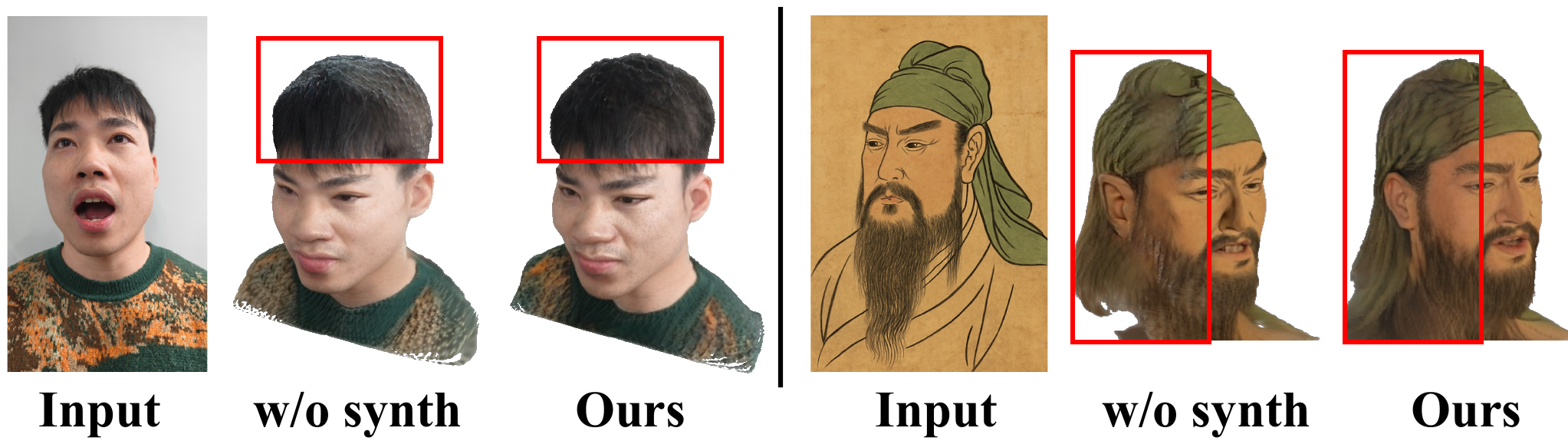}
    \caption{\textbf{Visual analysis of 3D consistency under varying training data settings.}}
    \label{fig:rebuttal_synth}
\end{figure}

\begin{table}[t]
\centering
\scalebox{0.85}{
\begin{tabular}{cccc}
\toprule
\textbf{Real / Synthetic Datasets} & \textbf{NeRSemble-v2} & \textbf{Ours} & \textbf{CAP4D} \\ \hline
\textbf{Spatial WE ($\times 10^{-2}$) $\downarrow$} & 2.377  &  4.252  &  10.45  \\
\textbf{Temporal WE ($\times 10^{-4}$) $\downarrow$}  & 4.605  &  7.868  &  31.27 \\
\bottomrule
\end{tabular}
}
\caption{\textbf{Data quality evaluation.} We use Warping Error (WE) as metric to evaluate spatial and temporal consistency of datasets.}
\label{tab:dataset_eval}
\end{table}

\begin{table*}[t]
\centering
\scalebox{0.81}{
\begin{tabular}{c|ccc|ccc|c}
\toprule
\textbf{Method} & \textbf{InvertAvatar~\cite{zhao2024invertavatar}} & \textbf{DiffusionRig~\cite{ding2023diffusionrig}} & \textbf{GPAvatar~\cite{chu2024gpavatar}} & \textbf{LAM~\cite{he2025lam}} & \textbf{GAGAvatar~\cite{chu2024gagavatar}} & \textbf{Portrait4D-v2~\cite{deng2024portrait4dv2}} & \textbf{Ours} \\ \hline
\textbf{Render Quality $\uparrow$}         & 1.4  & 1.95 & 2.7  & 2.51 & 3.29 & 3.48 & \textbf{4.37}  \\
\textbf{Motion Consistency $\uparrow$}     & 1.85 & 2.05 & 2.73 & 2.93 & 3.45 & 3.4  & \textbf{4.17}  \\
\textbf{Identity Preservation $\uparrow$}  & 1.7  & 2.15 & 2.74 & 2.54 & 3.44 & 3.54 & \textbf{4.23}  \\
\bottomrule
\end{tabular}
}
\caption{\textbf{User study evaluation.} We ask users to rate the results in 1-5, the higher the better.}
\label{tab:user_study}
\end{table*}

\section{Synthetic Dataset}
\label{sec:synth_suppl}

\begin{figure*}[t]
    \centering
    \includegraphics[width=1.0\textwidth]{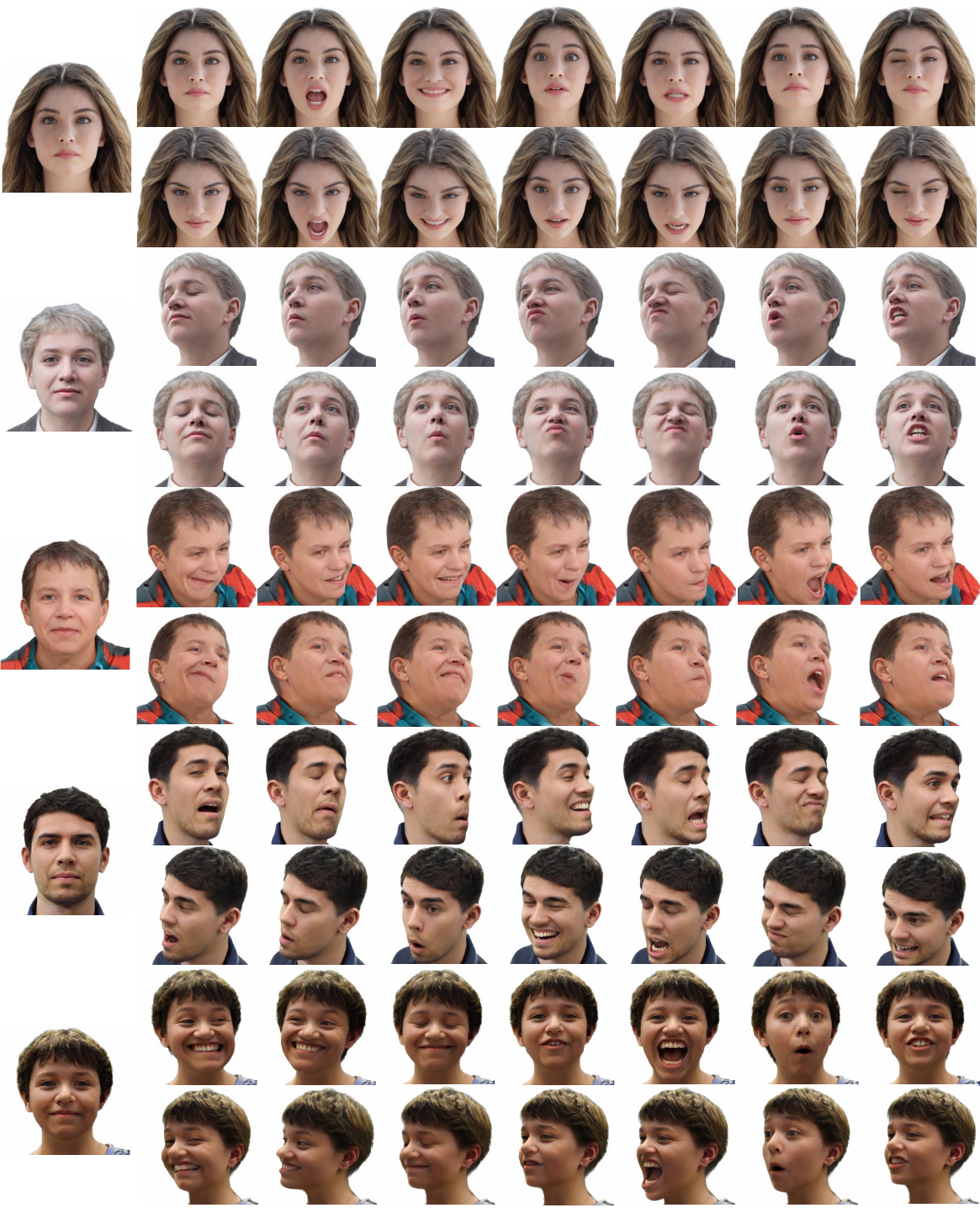}
    \caption{\textbf{Visualization of examples from our synthetic dataset.}}
    \label{fig:synth_show}
\end{figure*}

\noindent\textbf{Visualization.} In Sec.~\textcolor{cvprblue}{3.4} of the main paper, we explain the curation process of our synthetic multi-view head dataset. In this section, we provide visualization results of this dataset, as shown in Fig.~\ref{fig:synth_show}, which illustrates the results of each identity under different camera viewpoints and expressions.

\noindent\textbf{Quality assessment of our dataset.} We evaluate our synthetic dataset in Tab.~\ref{tab:dataset_eval} by using warping error as in HuGe100K~\cite{zhuang2024idolinstant}. Our generated dataset achieves numeric results comparable to the real-captured dataset NeRSemble~\cite{kirschstein2023nersemble} and outperforms a synthetic dataset using CAP4D~\cite{taubner2025cap4d} in both spatial and temporal dimensions.

Our synthetic data achieves a well‑balanced combination of identity diversity and expression richness, while maintaining multi‑view and 3D consistency. Such a dataset contributes to training a more robust model. Please refer to our supplementary video for additional dynamic results.

\section{Additional Comparison Results}
\label{sec:comp_suppl}

\begin{figure*}[t]
    \centering
    \includegraphics[width=0.9\textwidth]{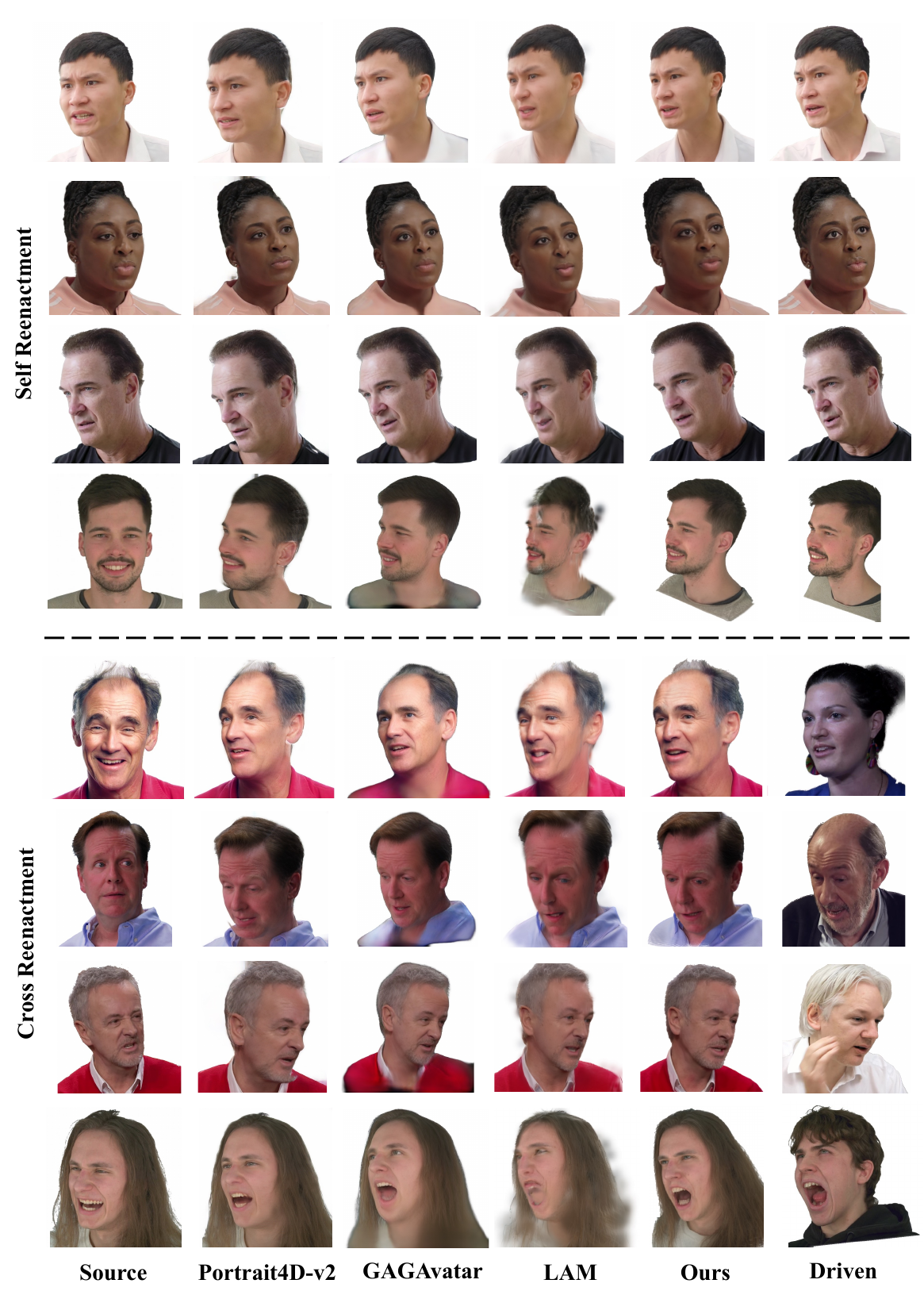}
    \caption{\textbf{Visualization of self and cross reenacted results on the VFHQ and NeRSemble-v2 datasets for the monocular input setting.}}
    \label{fig:mono_comp_suppl}
\end{figure*}

\noindent \textbf{Monocular Setting.} In Fig.~\ref{fig:mono_comp_suppl}, we show more self and cross reenactment results on the VFHQ dataset and NeRSemble-v2 dataset.

\begin{figure*}[t]
    \centering
    \includegraphics[width=1.0\textwidth]{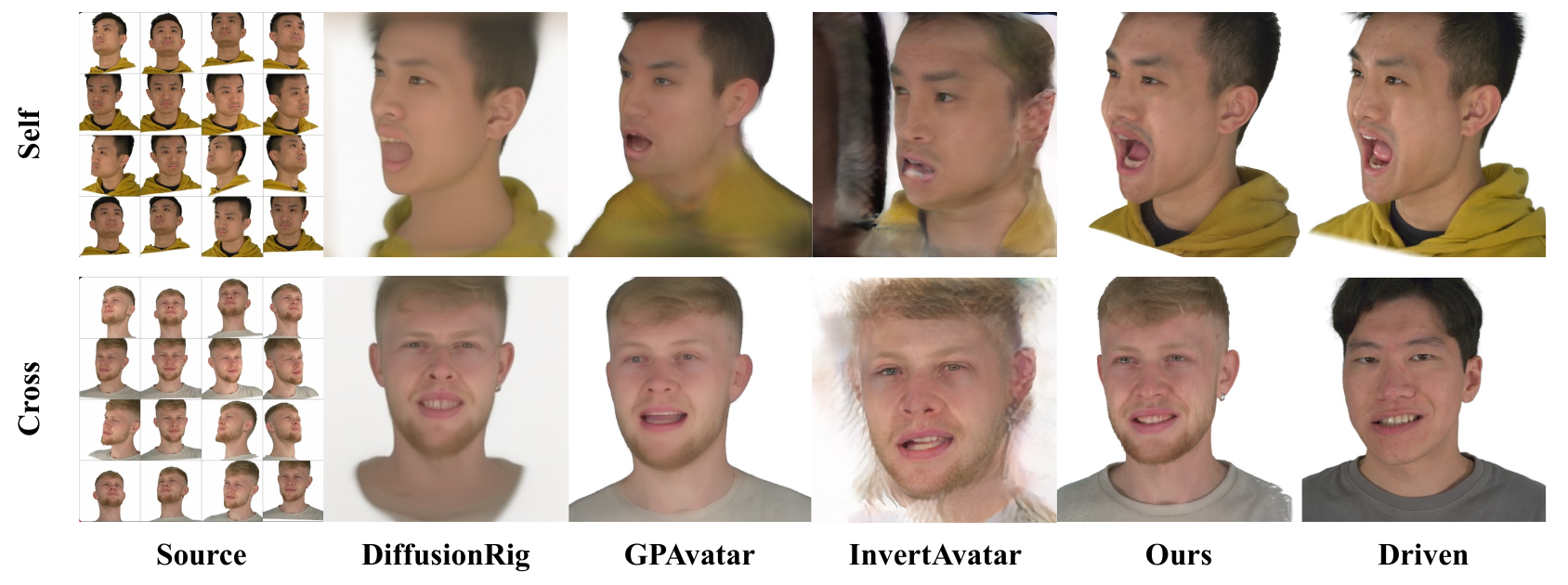}
    \caption{\textbf{Visualization of self and cross reenacted results on the NeRSemble-v2 dataset for the multi-view setting.}}
    \label{fig:mv_comp_suppl}
\end{figure*}

\noindent \textbf{Multi-view Setting.} In Fig.~\ref{fig:mv_comp_suppl}, we show more results of self and cross reenactment on the NeRSemble-v2 dataset. Please refer to our supplementary video for additional dynamic results.

\noindent \textbf{User Study.} We have also included a human evaluation in Tab.~\ref{tab:user_study} as an additional validation. Our method outperforms baselines in render quality, motion consistency, and identity preservation.

\section{Additional Ablation Results}
\label{sec:abla_suppl}

\subsection{Ablation on training data}
\label{subsec:abla_data_suppl}

Thanks to the paradigm of our framework, the model can accept an arbitrary number of input images. Although the number of input views during training is limited to $1 \sim 16$ due to VRAM constraints, similar to VGGT~\cite{wang2025vggt}, our model can take more than 16 input images during inference. This flexibility enables us to train on monocular video datasets, unlike methods such as Avat3r~\cite{kirschstein2025avat3r} that require a fixed set of four input views and therefore rely exclusively on multi‑view datasets. The monocular video dataset VFHQ~\cite{xie2022vfhq} contains approximately 7k identities, which is an order of magnitude larger than existing multi‑view datasets such as NeRSemble‑v2~\cite{kirschstein2023nersemble}, Ava‑256~\cite{ava256}, and RenderMe-360~\cite{pan2024renderme360}, each of which typically includes only a few hundred identities.

To evaluate the importance of high-quality training data, we prepare two ablated versions.
One model is only trained on the NeRSemble-v2 dataset, as shown in Fig.~\ref{fig:abla_data_suppl} (b), which can hardly preserve the identity of input images.
When using both NeRSemble-v2 and a rich-identity dataset VFHQ, the model generalizes better to novel identities, but would collapse in some extreme viewpoint in Fig.~\ref{fig:abla_data_suppl} (c).
When including our multi-view synthetic data, our model demonstrates superior generalization capability of identity as shown in Fig.~\ref{fig:abla_data_suppl} (d).
As shown in Fig.~\ref{fig:rebuttal_synth}, our model achieves better 3D consistency using our synthetic dataset.

\subsection{Ablation on our method}
\label{subsec:abla_method_suppl}

\begin{figure*}[t]
    \centering
    \includegraphics[width=1.0\textwidth]{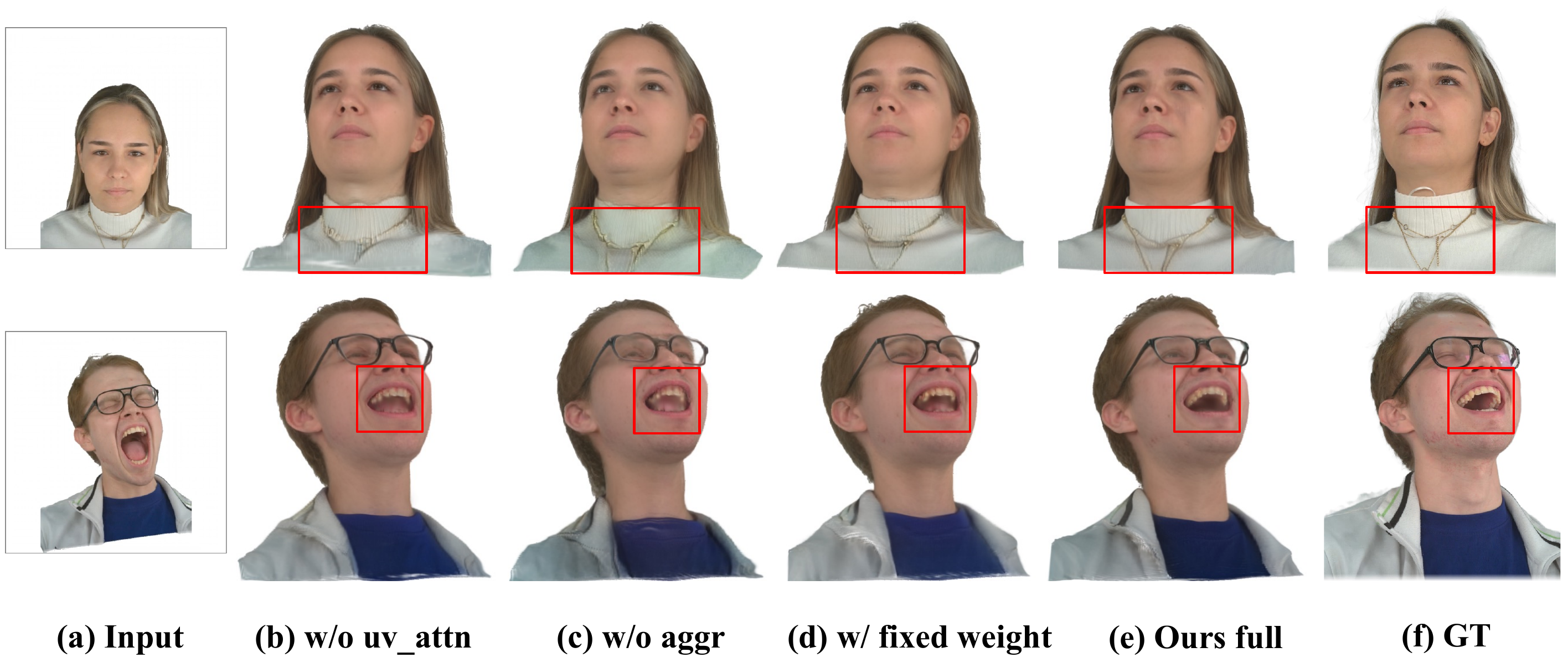}
    \caption{\textbf{Visualization of ablation study results of our method.}}
    \label{fig:abla_method_suppl}
\end{figure*}

In this section, we provide additional visualizations of ablation studies on our method. 
Other than the ablated versions in the main paper, we further include an extra ablation on our self‑adaptive fusion strategy, as shown in Fig.~\ref{fig:abla_method_suppl} (d). 
In our full model, the fusion weight for each Gaussian is predicted by the network as a per‑Gaussian value in the range $[0, 1]$. In contrast, this ablated variant replaces the learned weight with a fixed value computed as \(0.5\) times the UV-domain confidence map described in Sec.~\textcolor{cvprblue}{3.1}.
The results demonstrate that our proposed full model effectively leverages information from the input views, leading to higher‑fidelity head avatar reconstruction. Please refer to our supplementary video for additional dynamic results.

\section{Applications}
\label{sec:app_suppl}

\begin{figure*}[t]
    \centering
    \includegraphics[width=1.0\textwidth]{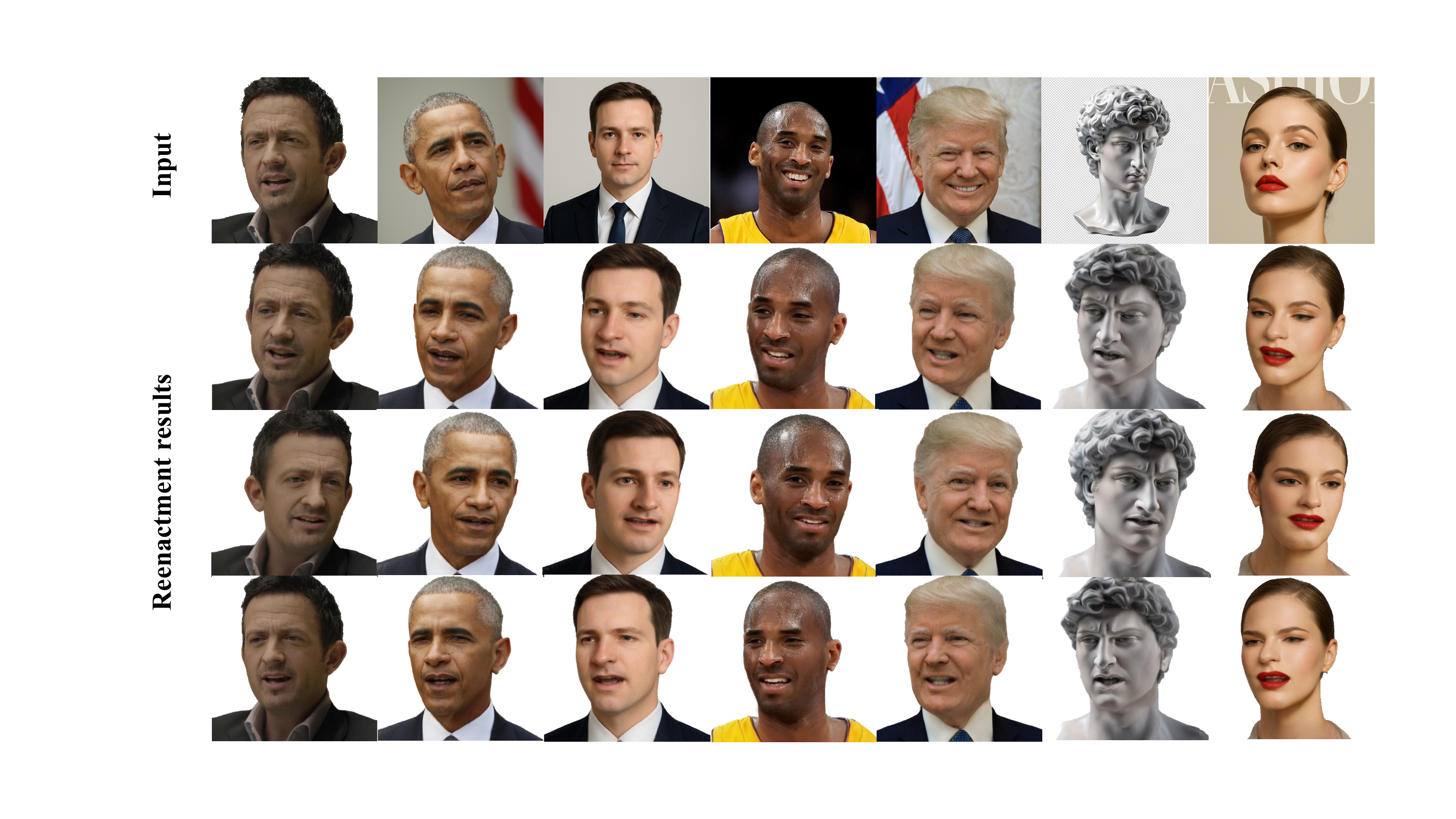}
    \caption{\textbf{Visualization of in-the-wild cases.}}
    \label{fig:apps_image}
\end{figure*}

\noindent\textbf{In-the-wild Image Reenactment.} We also demonstrate the reenactment results of our method on in‑the‑wild Internet cases, as shown in Fig.~\ref{fig:apps_image}.

\begin{figure*}[t]
    \centering
    \includegraphics[width=1.0\textwidth]{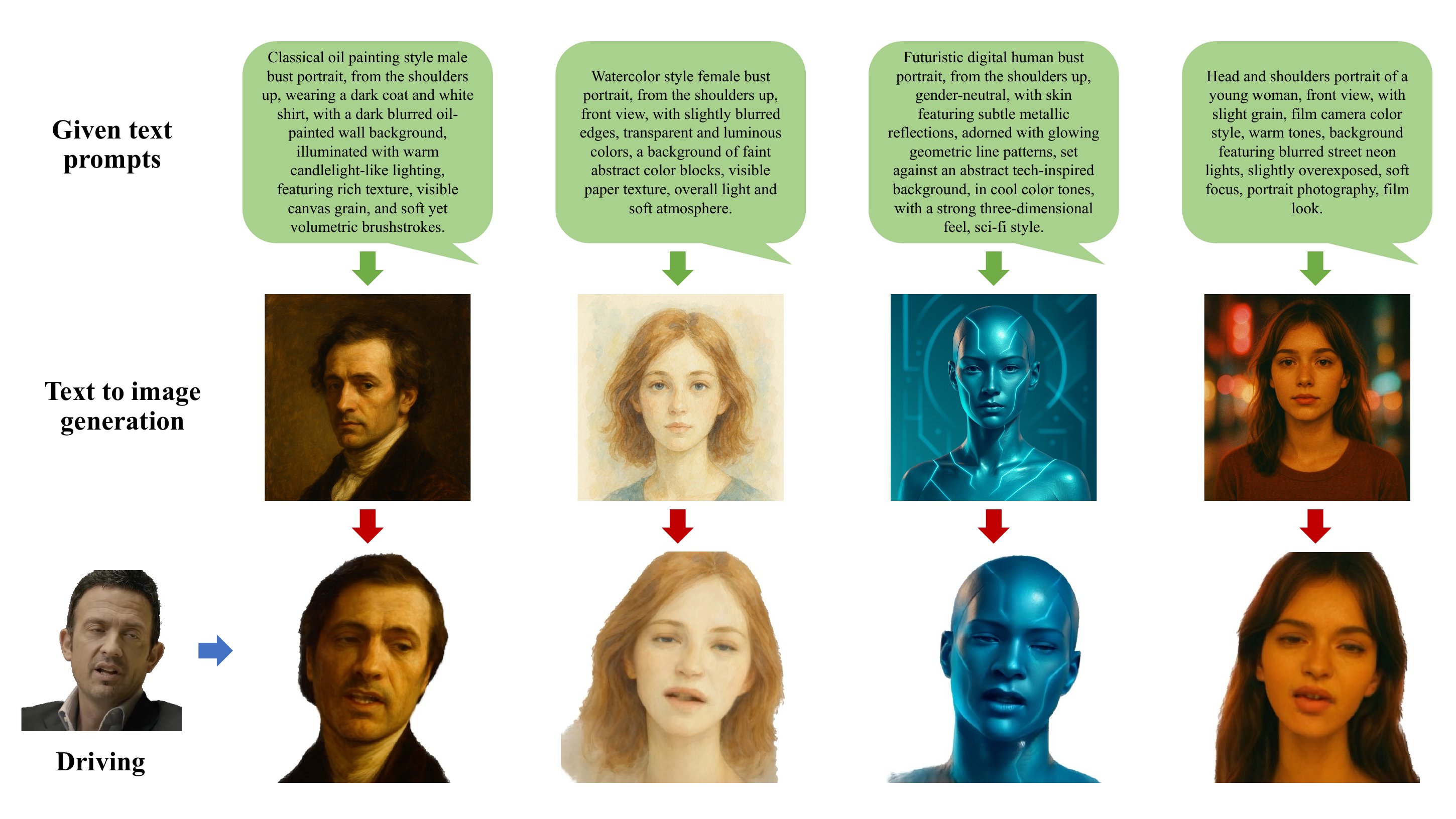}
    \caption{\textbf{Visualization of text-to-head-avatar generation.}}
    \label{fig:apps_text}
\end{figure*}

\noindent\textbf{Text-to-Head-Avatar Generation.} In addition, we visualize the pipeline for generating controllable head avatars from text prompts. Given a textual description, we employ advanced multimodal large models such as ChatGPT or Gemini to synthesize corresponding images, which are then fed into our model to produce a animatable head avatar. Detailed visualizations are provided in Fig.~\ref{fig:apps_text}.

Such results show that our method generalizes well to a wide variety of visual styles, benefiting from both our proposed approach and the synthetic dataset. Please refer to our supplementary video for additional dynamic results.

\section{Limitations}
\label{sec:limit}

Despite its effectiveness, our approach has several limitations. First, the expressiveness of our reconstructed head avatars is inherently constrained by the FLAME~\cite{sigasia17FLAME} model used for both data tracking and avatar driving. As a result, fine‑grained facial dynamics such as subtle wrinkles, micro‑expressions, and tongue motions cannot be reliably captured or reproduced. Second, although our training includes both real and synthetic data, the combined dataset still exhibits certain demographic biases, which may lead to degraded performance or failure cases for under-represented groups. Third, while our framework supports an arbitrary number of input images, the computational cost and memory consumption grow with the number of views, whereas the performance improvement saturates beyond a certain point. These limitations highlight important directions for future work, such as integrating more expressive parametric models, reducing data bias, and improving scalability for large‑view inference.

\section{Ethics}
\label{sec:ethics}

Our work focuses on feed‑forward reconstruction of animatable head avatars from arbitrary numbers of input facial images. While the proposed method advances the efficiency and accessibility of personalized head avatar creation, it also raises several potential ethical concerns. First, the ability to reconstruct high‑fidelity 3D human heads from sparse or casually captured images introduces risks of misuse, such as generating unauthorized digital replicas of individuals or producing manipulated content that may compromise privacy, consent, or identity integrity. Second, reconstructed avatars could be misappropriated for malicious applications, including impersonation, deepfake‑style synthesis, or other forms of deceptive media generation.

To mitigate these risks, our research uses only publicly available datasets with established licenses and synthetic data generated in‑house. We emphasize that our method is intended for legitimate applications such as virtual telepresence, animation, and human computer interaction. We strongly discourage any use of this technology for surveillance, non‑consensual persona reproduction, or deceptive content creation. Future deployment of systems built upon our approach should incorporate suitable safeguards, such as perceptual watermarking, provenance tracking, or identity verification mechanisms, to ensure responsible and ethical use.

\clearpage

\end{document}